\documentclass[lettersize,journal]{IEEEtran}
\usepackage{amsmath,amsfonts}
\usepackage{float}
\usepackage{algorithm,algpseudocode}
\usepackage{array}
\usepackage[caption=false,font=normalsize,labelfont=sf,textfont=sf]{subfig}
\usepackage{textcomp}
\usepackage{stfloats}
\usepackage{url}
\usepackage{verbatim}
\usepackage{supertabular}
\usepackage{graphicx}
\usepackage{cite}
\def\BibTeX{{\rm B\kern-.05em{\sc i\kern-.025em b}\kern-.08em
    T\kern-.1667em\lower.7ex\hbox{E}\kern-.125emX}}
    
\usepackage{cite}
\usepackage{amsmath,amssymb,amsfonts}
\usepackage{algpseudocode}
\usepackage{lipsum}
\usepackage{graphicx}
\usepackage{textcomp}
\usepackage{xcolor}
\usepackage{tabularx}
\usepackage{longtable}
\usepackage{supertabular}
\usepackage{adjustbox}
\usepackage[font=footnotesize]{caption}
\usepackage{svg}

\usepackage{lastpage}
\usepackage{fancyhdr}
\def\BibTeX{{\rm B\kern-.05em{\sc i\kern-.025em b}\kern-.08em
    T\kern-.1667em\lower.7ex\hbox{E}\kern-.125emX}}
    
\usepackage{flushend}

\usepackage{todonotes}
\usepackage{comment}

\begin{document}
    
\title{A 1.6-fJ/Spike Subthreshold Analog \\ Spiking Neuron in 28 nm CMOS  }

\author{

{Marwan Besrour \IEEEauthorrefmark{1}, Jacob Lavoie \IEEEauthorrefmark{1}, Takwa Omrani \IEEEauthorrefmark{1}, Samuel Bouchard \IEEEauthorrefmark{1}, Gabriel Martin-Hardy \IEEEauthorrefmark{1}, Esmaeil Koleibi \IEEEauthorrefmark{1}, Jeremy Menard \IEEEauthorrefmark{1}, Konin Koua \IEEEauthorrefmark{1}, Philippe Marcoux \IEEEauthorrefmark{1}, Mounir Boukadoum \IEEEauthorrefmark{2} , Rejean Fontaine \IEEEauthorrefmark{1} }

\IEEEauthorblockA{
\IEEEauthorrefmark{1} Department of Electrical Engineering and Computer Science\\ 
\IEEEauthorrefmark{1} 3IT Interdisciplinary Institute for Technological Innovation, Université de Sherbrooke, Canada\\
\IEEEauthorrefmark{2} Département d'informatique, Université du Québec à Montréal, Canada\\
Marwan.Besrour@Usherbrooke.ca 
}   
        
\thanks{This paper was produced by the IEEE Publication Technology Group.}
\thanks{Manuscript received April XX, 202X; revised August XX, 202X.}}




\maketitle

\begin{abstract}

The computational complexity of deep learning algorithms has given rise to significant speed and memory challenges for the execution hardware. In energy-limited portable devices, highly efficient processing platforms are indispensable for reproducing the efficiency achieved by much bulkier platforms. 

In this work, we present a low-power Leaky Integrate-and-Fire (LIF) neuron design fabricated in TSMC's 28 nm CMOS technology as proof of concept to build an energy-efficient mixed-signal Neuromorphic System-on-Chip (NeuroSoC). The fabricated neuron consumes 1.61 fJ/spike and occupies an active area of 34 $\mu m^{2}$, leading to a maximum spiking frequency of 300 kHz at 250 mV power supply.

These performances are used in a software model to emulate the dynamics of a Spiking Neural Network (SNN). Employing supervised backpropagation and a surrogate gradient technique, the resulting accuracy on the MNIST dataset, using 4-bit post-training quantization stands at 82.5\%. The approach underscores the potential of such ASIC implementation of quantized SNNs to deliver high-performance, energy-efficient solutions to various embedded machine-learning applications.

\end{abstract}


\section{Introduction}
\IEEEPARstart{B}{}iology-inspired neuron cells are the fundamental computing units of neuromorphic systems \cite{bentivoglio1999history}, and their energy efficiency is closely linked to them, highlighting the importance of the neuron model choice. In 1943, \cite{Palm1986} proposed the first artificial neuron model, based on binary logic principles. They were followed in 1952 by \cite{hodgkin1939action} who developed a continuous, conductance-based, model that more accurately described the biological neuron activity.

\cite{Palm1986} model was closely linked to the digital computing paradigm with 1's and 0's, rendering it easier to simulate on digital computers than the differential equation-based model proposed by \cite{hodgkin1939action}, but the inter-neural connection weights of the model were manually set, severely limiting its learning capacity. In the sixties, Mead noticed that the differential equations governing the dynamics of biological neurons bore a close resemblance to those governing the dynamics of CMOS transistors \cite{Mead1989}. Therefore, a few transistors operating in the analog domain could replace thousands of transistors working collectively on a digital chip to emulate differential equations.

Artificial spiking neurons aim to emulate the characteristics of biological neurons while maximizing the computational capabilities, flexibility and robustness of the neuromorphic systems that use them. Spiking neurons differ from the former by their intermittent spiking feature instead of continuous output, which leads to substantial energy saving. The mathematical models are typically implemented using CMOS transistors, with each functionality incurring costs in terms of silicon footprint and power dissipation \cite{Indiveri2011Neuromorphic}. The key performance metrics of the resulting CMOS circuits consist of several parameters such as total energy consumption, chip area and biological plausibility.

Energy consumption is estimated both in terms of static power consumption and dynamic energy per spike. Additional metrics include supply voltage, mean and maximum spiking frequency, spiking and bursting patterns, runtime reconfigurability (thresholds, time constants, refractory period), spike frequency adaptation, intrinsic plasticity, load driving capability, synaptic integration with non-volatile memory (NVM) devices such as memristors, and robustness against Process Variation and Temperature (PVT) variations. Achieving the right balance of these parameters is crucial for building efficient neuromorphic systems for the target application.

The system of differential equations in the Hodgkin and Huxley model of a biophysical neuron is difficult to simulate on a large scale. In 2003, \cite{Izhikevich2003Tran,Izhikevich2004} proposed a more efficient model in this respect, demonstrating an array of cortical spiking and bursting behaviors \cite{Izhikevich2003Tran,Izhikevich2004}. Early silicon implementations of the model include the works in \cite{WIJEKOON2008524,Rangan2010IZ,Schaik2010ISCAS}. More recent work includes an implementation predicated on a ring oscillator in 65 nm CMOS node \cite{Zhang2017neuron}. The spiking and bursting modes of this neuron \cite{Zhang2017neuron} were dictated by digital control signals, which can be supplied by an FPGA. The circuit was optimized with respect to area, energy consumption, and display resilience against PVT variations as corroborated in \cite{Zhang2018ISCAS}.

Another spiking model is proposed in  \cite{Morris1981VoltageOI}. The model is described by two nonlinear differential equations linked to exponential functions. There is a membrane potential equation that features instantaneous activation of Calcium current, coupled with an additional equation that details the slower activation of Potassium current.

In yet another work \cite{Sourikopoulos20174fj}, two neuron implementations are explained. A bio-inspired version, equipped with leak conductance, feedback loop, and related features, is proposed alongside a streamlined version that uses fewer transistors and enables a higher spiking rate. The energy expenditure per spike was evaluated at 4fJ/spike, with the spatial footprint of the simplified version determined to be 35 $\mu$m2. Subsequently, the proposed bio-inspired version was optimized on 55nm CMOS node in \cite{Ferreira2019}. The first implementation, consuming a mere 2.6 fJ/spike, exhibited Fast Spiking, while the Low Threshold Spiking (LTS) second neuron with additional components, demonstrated spike frequency adaptation.

An integrator circuit at the input can generate exponential sub-threshold behavior from a steady input current and mimic leak conductance. The current-mode low-pass filter methodologies presented in academic works include Differential Pair Integrator (DPI), Tau Cell, and Low-pass filter \cite{Chicca2014NeuromorphicEC}. The DPI is the most condensed and least intricate implementation. The crucial elements of such a neuron include a positive feedback circuit that simulates sodium channel activation and potassium channel deactivation for prompt toggling, a block for spike frequency adaptation, a comparator, a block to generate spikes, and a circuit for reset and refractory period.

The adaptive exponential Integrate and Fire neuron, also known as I\&F neuron, includes spike frequency adaptation. Essentially, the neuron adjusts to repeated stimuli, leading to a rise in the voltage threshold and a subsequent decrease in the neuron's firing rate. \cite{Boahen1998RetinomorphicVS} demonstrated the feasibility of spike-frequency adaptation by linking a four-transistor "current-mirror integrator" in negative-feedback mode to any Integrate and Fire circuit. This idea was further adopted by \cite{Indiveri2003ISCAS}, where the implementation of AdexIF neuron was showcased. 

This neuron concept underwent evolution over time, with feature size reduction and the incorporation of attributes such as threshold voltage and refractory period modulation. One implementation is featured in \cite{Rubino2019UltraLowPS} and is fabricated in 22 nm CMOS FDSOI technology.

The leaky-integrate-and-fire modeling approach offers a fair compromise between computational complexity and biological features. 
The LIF is the simplest model to emulate a biological neuron, as it only accounts for the leaky integration and the spiking behaviour. Although many works have been published about modeling the spiking behavior of biological neurons, the LIF model is a generally accepted  trade-off between the complexity of biologically plausible neuron models \cite{gerstner2014neuronal}. The simplicity of this model makes it promising for the implementation of large-scale NeuroSoC processors in real-life applications, as is the case for the IBM TrueNorth chip \cite{TrueNorth2014} and Intel Loihi \cite{Loihi}. However, they are implemented with digital architectures. 

In recent years, several analog CMOS implementations of the LIF neuron have explored the trade‐offs among energy per spike, silicon area, and supply voltage. Indiveri \textit{et al.} demonstrated a 0.35 µm analog LIF neuron consuming on the order of 900 pJ/spike with a 2 573 $\mu m^2$ area at 3.3 V \cite{indiveri_vlsi_2006}, while Wijekoon and Dudek reported 9 pJ/spike in 2 800 $\mu m^2$ at 3.3 V in the same technology node \cite{wijekoon_compact_2008}. In 65 nm, \cite{Zhang2017neuron} implemented an Izhikevich neuron based on a ring oscillator achieving 58.5 fJ/spike in 472 $\mu m^2$ at 200 mV, and \cite{Sourikopoulos20174fj} presented a bio-inspired Moris-Lecar design consuming 4 fJ/spike in 35 $\mu m^2$ at 200 mV. Ferreira further optimized a streamlined LIF in 55 nm to 3.6 fJ/spike and 85 $\mu m^2$ at ±100 mV \cite{Ferreira2019}. These implementations illustrate that scaling and deep sub-threshold operation can dramatically reduce dynamic energy, yet a clear tension remains between achieving both ultra-low energy and minimal area at low voltages.

The primary focus of this work is the design and validation of a scalable Leaky Integrate-and-Fire (LIF) neuron architecture for CMOS implementation. While considerations such as communication protocols (e.g., I2C or SPI) are essential for practical deployments, they fall beyond the scope of this study and will be addressed in future work. Despite rapid advances in digital spiking‐neuroprocessor architectures, there remains an acute need for ultra‐low-power, high-density analog neuron building blocks that can operate at sub-threshold voltages (<300 mV) and sub-femtojoule energy per event to enable energy-constrained edge AI and biomedical implants. Existing analog LIF implementations either consume more energy or occupy larger areas when scaled to advanced nodes, limiting their applicability in large-scale SNNs on portable and implantable platforms. Our work addresses this gap by presenting the first 28 nm CMOS analog LIF neuron that simultaneously achieves 1.61 fJ/spike, a 34 $\mu m^2$ footprint, and 250 mV supply operation. We further validate the design through post‐silicon measurements on 20 fabricated ASICs and develop a hardware-aware software model complete with co-simulation measurement of f-I curves and surrogate-gradient SNN training to demonstrate end-to-end feasibility for edge applications.

This paper explores the modeling, behavior, and energy consumption of an analog LIF neuron implemented in CMOS 28 nm technology. In Section II, the paper introduces the architecture of the LIF model and its CMOS implementation. Section III focuses on the validation and characterization of the LIF neuron design. Section IV presents the results obtained from measurements of 20 ASIC samples. Additionally, it includes a software model to emulate the analog LIF neuron, which is utilized to train a deep SNN on the MNIST dataset. This section also evaluates the energy per prediction figure of merit. Finally, Section V concludes the research presented in this paper.

\section{Materials}

\subsection{LIF Model}

The Leaky Integrate-and-Fire (LIF) model is based on the observation that a biological neuron essentially integrates input currents and generates output spikes (action potentials) when the membrane voltage level crosses a predefined threshold. The neuron receives an input current $I_\text{syn}$ from presynaptic neurons through synapses, represented by weighted connections. $I_\text{syn}$ increases the electrical charge inside the cell, with the cell's membrane acting like a leaky capacitor with respect to the outside, and a resistance in parallel setting the leakage current for a given membrane potential $V_\text{mem}$. The latter is set by the following differential equation:

\begin{equation}
     C_{mem}\frac{dV_\text{mem}(t)}{dt} = I_\text{syn}(t) - \frac{dV_\text{mem}(t)}{R_m}
     \label{equ:V_m} 
\end{equation}

In practice, equation \ref{equ:V_m} is discretized to enable numerical simulation. Using a time step ($dt$) and a resting membrane potential $V_\text{reset}$, it becomes:

\begin{equation}
V_\text{mem}(t+1) = V_\text{mem}(t) + \frac{-(V_\text{mem}(t) - V_\text{reset}) + R_{m} I_\text{syn}(t)}{\tau_{m}}\cdot dt
\end{equation}

This equation represents a balance between two forces: the leaky term $-(V_\text{mem}(t) - V_\text{reset})$, which drives the membrane potential towards the reset potential, and the input current term $R_m I_\text{syn}(t)$, which pushes the membrane potential in the opposite direction. 

When $V_\text{mem}(t)$ reaches the threshold potential $V_\text{th}$, the neuron generates an output spike and the membrane potential returns to $V_\text{reset}$. Finally, to account for the refractory period of biological neurons, it is prevented from generating new spikes during $t_\text{ref}$. 

The parameters that govern the LIF neuron's operation are critical for the accurate emulation of neuronal dynamics. A clear understanding of these parameters is necessary to allow tuning the model to reflect the behaviors observed in biological systems. 

\begin{itemize}
    \item $V_\text{reset}$: the voltage to which the membrane potential ($V_\text{mem}$) returns after a spike is generated.
    \item $V_\text{th}$: the membrane voltage at which the neuron fires a spike.
    \item $V_\text{spike}$: the LIF output when it generates a spike. 
    \item $\tau_m$: the time constant of the membrane potential response to input currents. It is the product of the membrane resistance ($R_{m}$) and the membrane capacitance ($C_{mem}$).
    \item $dt$: the discrete time interval to update the membrane potential of the model. Smaller values result in more accurate simulations at the cost of increased computational complexity.
\end{itemize}

\subsection{LIF Model Computation Sequence}

Algorithm \ref{Algo1} below provides the pseudo-code of the LIF model's operation. It computes the membrane potential \(V_{\text{mem}}\) of a neuron over time, integrating incoming synaptic currents (\(I_{\text{syn}}(t)\)) and producing spikes when the potential crosses a certain threshold (\(V_{\text{th}}\)). Initially, \(V_{\text{mem}}\) is set to a reset value \(V_{\text{reset}}\). Then, the model updates \(V_{\text{mem}}\) at each time step (\(dt\)) based on the synaptic input (\(I_{\text{syn}}(t)\)), the membrane time constant (\(\tau_m\)) and the membrane resistance (\(R_m\)). If \(V_{\text{mem}}\) exceeds \(V_{\text{th}}\), it is reset to \(V_{\text{reset}}\), simulating the firing of a spike, and \(V_{\text{spike}}\) is output. The process incorporates a refractory period (\(t_{\text{ref}}\)), during which the neuron cannot fire again. This model captures the essential dynamics of neuronal spiking behavior using mathematical and logical constructs.

\begin{algorithm}
\caption{Leaky Integrate-and-Fire (LIF) Model}
\label{Algo1}
\begin{algorithmic}
\Require $V_\text{reset}$, $V_\text{th}$, $V_\text{spike}$, $\tau_m $, $R_m$, $dt$, $t_\text{ref}$
\State $V_\text{mem} \leftarrow V_\text{reset}$
\For{each time step $dt$}
    \State Calculate $I_\text{syn}(t)$ as a sum of input currents from presynaptic neurons
    \If{$(t - t_\text{last\_spike}) \geq t_\text{ref}$}
        \State $V_\text{mem}(t+1) \leftarrow V_\text{mem}(t) + dt * (-(V_\text{mem}(t) - V_\text{reset}) + R_m * I_\text{syn}(t)) / \tau_m$
        \If{$V_\text{mem}(t+1) \geq V_\text{th}$}
            \State $V_\text{mem}(t+1) \leftarrow V_\text{reset}$
            \State Output spike = $V_\text{spike}$
            \State $t_\text{last\_spike} \leftarrow t$
        \Else
            \State Output spike = 0
        \EndIf
    \EndIf
\EndFor
\end{algorithmic}
\end{algorithm}

\subsection{LIF Implementation in CMOS}

\begin{figure}[h]
\centerline{\includegraphics[scale=0.2]{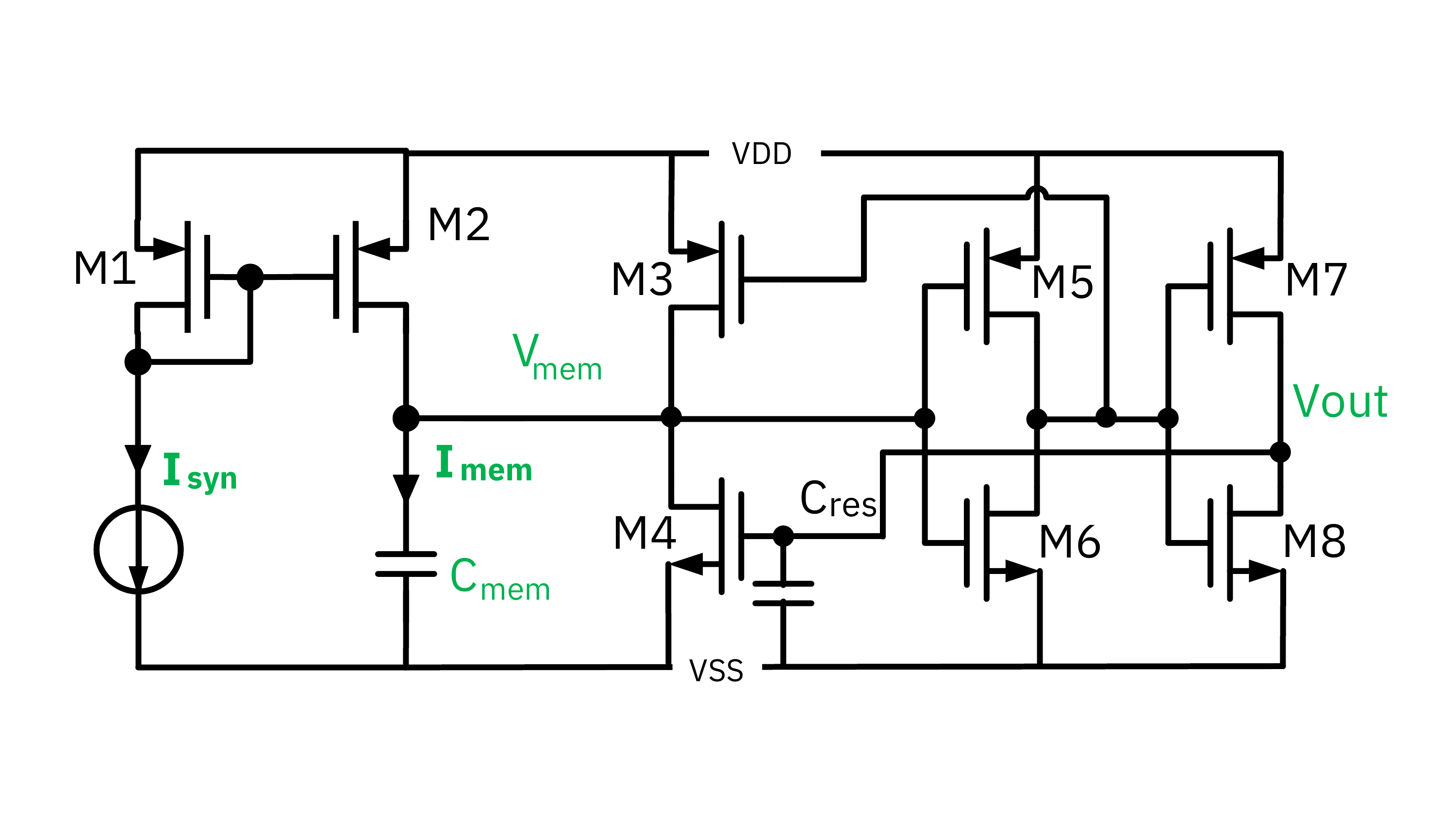}}
\caption{Proposed neuron circuit in CMOS. $M_{1}$-$M_{2}$-$C_{mem}$: Integration block; $M_{3}$-$M_{4}$-$C_{res}$: Reset block; $M_{5}$-$M_{8}$: Spike generation block.} 
\label{fig:neuron_architecture}
\end{figure}

Fig \ref{fig:neuron_architecture} provides the schematic of the LIF neuron circuit in CMOS, showing the Integration block ($M_{1}$-$M_{2}$-$C_{mem}$), the Reset block ($M_{3}$-$M_{4}$-$C_{res}$), and the Spike generation block ($M_{5}$-$M_{8}$). Each blocks is discussed next, focusing on the function of each transistor and the relationship with the mathematical model of the LIF.

\paragraph{Integration Block}
The Integration block consists of a current mirror made of transistors $M_1$ and $M_2$, and capacitor $C_{mem}$. The input current $I_{syn}$, which represents the neurotransmitter-induced current at the synapse between two neurons, flows through the drain of $M_1$ and a correlated current $I_D$ is generated at the drain of $M_2$. This current charges membrane capacitor $C_{mem}$ to develop a voltage $V_{mem}$ given by: 

\begin{equation}
V_{mem}(T)= \frac{1}{C_{mem}}\int_{T}^{}I_{D}~dt 
\label{eqn:eq3}
\end{equation}

Which yields:

\begin{equation}
V^{max}_{mem}(T)= \frac{I_{mem}\cdot T}{C_{mem}} + V_{reset}
\label{eqn:eq4}
\end{equation}

where $T$ is the integration period and $I_{mem} = I_{D}$. As the membrane voltage increases in absolute value, it approaches the threshold to trigger the spike generation block.

\paragraph{Spike Generation Block}
The Spike generation block comprises two back-to-back CMOS inverters that generate artificial action potentials at the neuron's output. The inverters consist of transistors $M_5$, $M_6$, $M_7$, and $M_8$. When $V_{mem}$ reaches the inversion threshold $V_{{th}}$, the inverters are activated producing a spike at the output. After producing a spike, the Reset block is activated to discharge the membrane capacitor $C_{mem}$ and prepare the neuron for the next cycle. If the inversion threshold is not reached and no incoming is integrated the leak pathway will slowly discharge the membrane capacitance.

The threshold voltage $V_{th}$ is a critical design parameter that determines when the membrane voltage $V_{mem}$ triggers the spike-generation inverter. A lower $V_{th}$ reduces the required integration time thus increasing the maximum firing frequency. Conversely, a higher $V_{th}$ improves robustness against noise and PVT variations but lengthens the inter-spike interval and marginally increases the average energy per spike due to the extended integration period. In our design, $V_{th}$ is set by the sub-threshold inverter pair M5-M6 and can be expressed as reported in \cite{SHARROUSH20181001}: 

\begin{equation}
V_{th} = \frac{V_{DD}}{2} + \frac{1}{2} \left( V_{tn} - |V_{tp}| \right) + n_n V_{T} \ln \left[ \sqrt{\frac{I_{0p}}{I_{0n}}}  \right]
\end{equation}

\paragraph{Reset Block}
The Reset block consists of transistors $M_3$ and $M_4$, and capacitor $C_{res}$. After a spike generation, this block is responsible for discharging membrane capacitor $C_{mem}$ and resetting membrane voltage $V_{mem}$ to its resting value $V_{reset}$. During the spike generation, $V_{out}$ increases across $C_{res}$ until it turns transistor $M_4$ on, thus discharging $C_{mem}$ and causing $V_{mem}$ to return to its resting potential, and $V_{out}$ to return to a low State. Then, the neuron is ready to start over the cycle of integration, comparison, and spike generation.

\paragraph{Leak Pathway}
The leak pathway in the proposed circuit is realized implicitly through the subthreshold operation of transistors M3 and M4. These transistors continuously discharge the integration capacitor, introducing a controlled leakage current that emulates the decay dynamics characteristic of the LIF model. This approach allows the circuit to maintain compliance with the LIF model's fundamental principles without the need for a dedicated leak pathway. The proposed circuit preserves the capability for spatio-temporal coincidence detection by leveraging the interplay between capacitive integration and the leakage current introduced by M3 and M4. Incoming synaptic currents are integrated and compared against the firing threshold within a temporal window determined by the leak dynamics. This enables the neuron to perform temporal filtering, ensuring that coincident inputs are effectively detected.

\subsubsection{Sub-Threshold Operation}

\paragraph{Drain Current}

Sub-threshold operation in MOS transistors occurs when the gate-to-source voltage ($V_{GS}$) is below the threshold voltage ($V_{TH}$), resulting in an exponential but weak increase in the drain current ($I_D$) with the slight change in $V_{GS}$ \cite{wang2006sub}. Hence, the drain current requires careful analysis to understand its impact on circuit behavior. In this particular mode, the drain current of a MOS transistor is given by  \cite{5395622,9856610}: 

\begin{equation}
I_D \approx I_{D}=I_0 \frac{W}{L} e^{\frac{V_{G S}-V_{TH}}{n v_{t}}}\left(1-e^{-\frac{V_{DS}}{v_{t}}}\right)
\label{equ:ID}
\end{equation}

Where $n$ and $I_{0}$ are technology dependent and $v_{t}=\frac{k_{B}T}{q}$ is the thermal voltage. What is peculiar in the weak inversion mode is the fact that the drain current is eponentially dependent on the gate-source volatge $V_{GS}$ and the drain-source voltage $V_{DS}$. To better highlight the dependence of the current on $V_{GS}$ and $V_{DS}$ we can rewrite the previous equation based on the analysis from \cite{5395622}: 

\begin{equation}
I_D = A \cdot e^{\frac{V_{G S}}{n v_t}} \cdot\left[e^{\frac{\lambda_{D S} V_{d S}}{n v_t}}\left(1-e^{-\frac{V_{D S}}{v_t}}\right)\right]
\end{equation}

Where A represent the stength of the channel inversion on the  zero-bias threshold voltage and the junction current going through source/drain to bulk:
\begin{equation}
A=I_0 \frac{W}{L} e^{-\frac{V_{T H 0}-\lambda_{B S} V_{B S}}{n v_t}}
\label{equ:A_exp}
\end{equation}

When $V_{DS}$ is relatively high or relatively low compared to $v_{t}$, equation Eq. (\ref{equ:ID_appx}) can be approximated as:
\begin{equation}
I_D \approx 
\begin{cases}
A \cdot e^{\frac{V_{G S}}{n v_t}} \quad \text { if } V_{D S} \gg v_t    \\
A \cdot e^{\frac{V_{G S}}{n v_t}} \cdot \frac{V_{D S}}{v_t} \quad \text { if } V_{D S} \ll v_t
\end{cases} 
\label{equ:ID_appx}
\end{equation}

From equation Eq. (\ref{equ:ID_appx}), we may rewrite $V_{SG1}$ under the first assumption as:
\begin{equation}
V_{SG1} = n v_t \cdot\ln{(\frac{I_{syn}}{A })}
\label{equ:A_exp}
\end{equation}

On the other hand, the drain current $I_{mem}$ in Fig. \ref{fig:neuron_architecture} can be expressed as:
\begin{equation}
I_{mem} =  A \cdot e^{\frac{V_{GS2}}{n v_t}}  \;\;\; \text { if } V_{DS2} \gg v_t
\end{equation}

Given that $V_{SG1} = V_{SG2}$ and under the assumption that $V_{D S} \gg v_t$, this yields:
\begin{equation}
I_{mem} =  A \cdot e^{\frac{n v_t \cdot\ln{(\frac{I_{syn}}{A })}}{n v_t}} = I_{syn}
\label{equ:Imem}
\end{equation}

which simply demonstrate that $I_{syn}$ is copied by $I_{mem}$ and integrated. Another way to interpret the previous simplifications is simply to consider the transistor as a lumped-circuit equivalent. In other words, when looking at the drain and source terminals, the MOS transistor is equivalent to a resistance $R_{syn}$ if $V_{D S} \ll v_t$ or a source current $I_{syn}$ if $V_{D S} \gg v_t$. Where $R_{syn}$ is the equivalent impedance seen from the drain of $M_{1}$ and given by:
\begin{equation}
R_{syn} =\frac{V_{D S}}{I_D}=\frac{v_t}{A \cdot e^{\frac{V_{G S}}{n v_t}}}
\label{equ:A_exp}
\end{equation}

Given we have $V_{SG1} = V_{DD} - R_{syn}I_{syn}$, we may rewrite $I_{mem}$ as:

\begin{equation}
I_{mem} =  A \cdot e^{\frac{V_{DD} - R_{syn}I_{syn}}{n v_t}} \cdot \frac{V_{DS2}}{v_t}  \;\;\; \text { if } V_{DS2} \ll v_t
\end{equation}

Replacin in Eq. \ref{eqn:eq4} yields $V^{max}_{mem}$ peak's voltage dependence on the input stimulation current $I_{syn}$ as follows:
\begin{equation}\label{eqn:eqVmem}
V^{max}_{mem}= \frac{A\cdot V_{DS2}}{v_t\cdot C_{mem}}  e^{\frac{V_{DD} - R_{syn}I_{syn}}{n v_t}} + V_{reset} 
\end{equation}

\paragraph{Output-High Voltage of Sub-Threshold Inverter}

Due to the sub-threshold operation of the transistors, the nominal high output voltage $V_{OH} = V_{DD} - \Delta V_{OH}$ of the CMOS inverter in the spike generation block, comprising cascaded CMOS inverters \(M_5\)-\(M_6\) and \(M_7\)-\(M_8\), does not reach \(V_{DD}\). The same goes for the nominal low output voltage $V_{OL} = \Delta V_{OL}$.  
Where both expressions are given by \cite{5395622,9856610}:

\begin{equation}
\Delta V_{OH} =v_t \frac{A_n}{A_p} e^{-\frac{V_{DD}}{n_p v_t}}
\label{eqn:eqVOH}
\end{equation}

\begin{equation}
\Delta V_{OL} =v_t \frac{A_p}{A_n} e^{-\frac{V_{DD}}{n_n v_t}}
\end{equation}

Both expressions yields the resulting voltage swing $V_{OH} - V_{OL}$ that deviates from $V_{DD}$:
\begin{equation}
\begin{aligned}
\Delta V & =\Delta V_{O H}+\Delta V_{O L} \\
& =v_t\left(\frac{A_n}{A_p} e^{-\frac{V_{D D}}{n_p v_t}}+\frac{A_p}{A_n} e^{-\frac{V_{D D}}{n_n v_t}}\right)
\end{aligned}
\end{equation}

From $\Delta V_{OH}$ we may conclude that the deviation of $V^{max}_{out}$ from $V_{DD}$ increases exponentially when the supply voltage decreases.

\subsection{ASIC implementation}

\begin{figure}[ht]
\centerline{\includegraphics[width=0.5\textwidth]{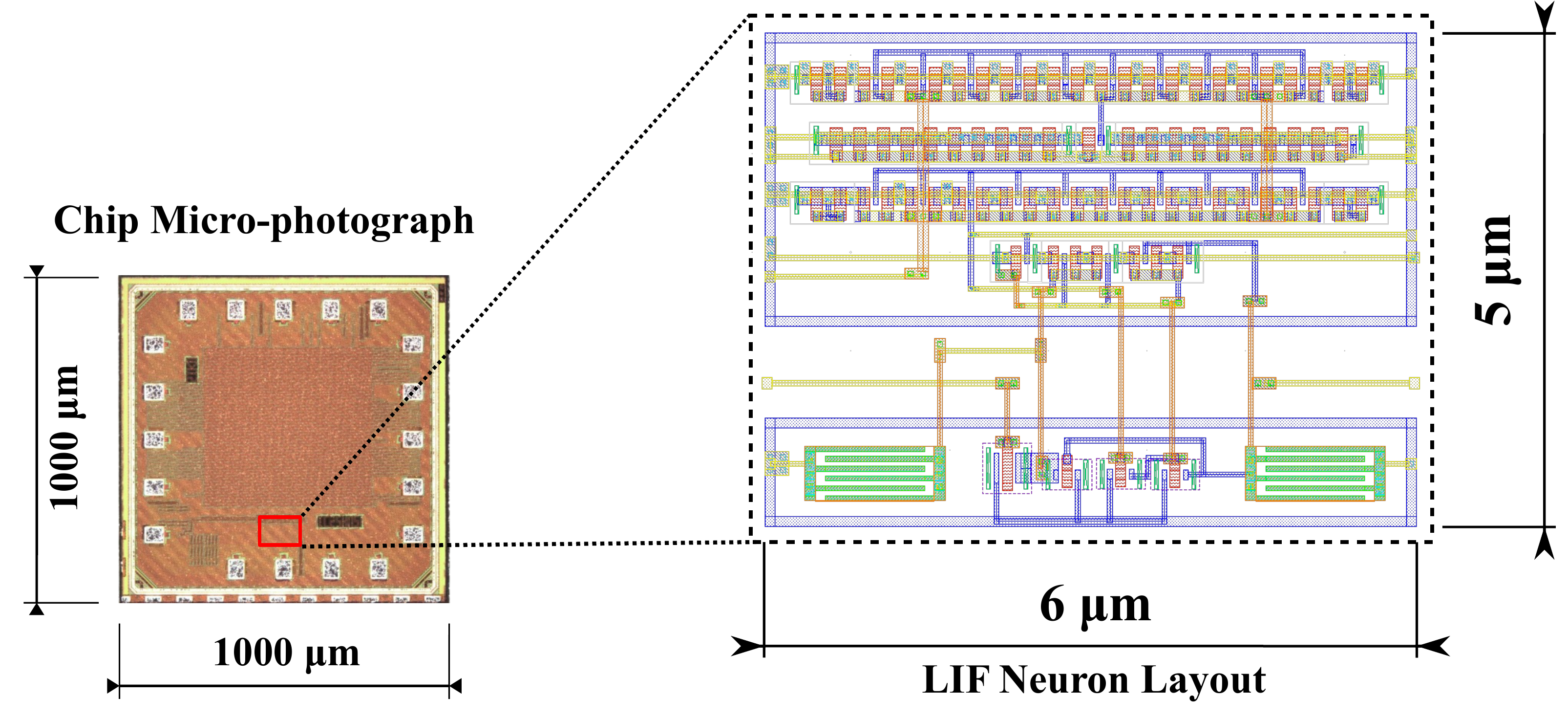}}
\caption{Layout implementation of the LIF neuron in TSMC 28 nm CMOS technology. The figure illustrates only the routing metals, excluding transistor-level details, due to non-disclosure agreements (NDAs) with the foundry. Post-layout simulations estimate gate-leakage currents to range from 100 fA to 1 pA under typical operating conditions, rendering them negligible compared to Isyn and Imem. The design complies with the manufacturer’s design rule checking (DRC) guidelines, including the minimum required lengths for gate polysilicon} 
\label{fig:ASIC}
\end{figure}

The proposed LIF neuron design was fabricated using the TSMC 28 nm CMOS technology node. The choice of the 28 nm TSMC CMOS technology for the silicon neuron design was driven by its unique advantages for neuromorphic circuit implementations. This technology node enables significant reductions in energy consumption and area footprint due to its smaller feature size and lower capacitance. Moreover, in smaller technology nodes such as 28 nm, transistors can be operated at lower core voltages, and logic gates can function effectively at reduced power supply voltages. These factors significantly lower overall power consumption, which is particularly advantageous for large-scale implementations in spiking neural networks. Additionally, the 28 nm process provides access to advanced features such as high-speed and low-leakage transistors, which are crucial for designing efficient and scalable neuromorphic circuits. These characteristics make the 28 nm technology particularly well-suited for the development of silicon neurons that can be integrated into large-scale spiking neural networks with minimal resource constraints

Fig. \ref{fig:ASIC} illustrates the circuit layout, which occupies an area of approximately $5~\mu m$ by $6~\mu m$. The total chip implementation has a total area of $1000~\mu m$ by $1000~\mu m$ and is encapsulated in a dual inline 20 package. To facilitate testing after fabrication, all the inputs, outputs, and power connections are linked to the ASIC's pads with Electro-Static Discharge (ESD) protection. 

\subsubsection{Validation and Characterization Test Bench}

\begin{figure}[hptb]
\centerline{\includegraphics[width=1\linewidth]{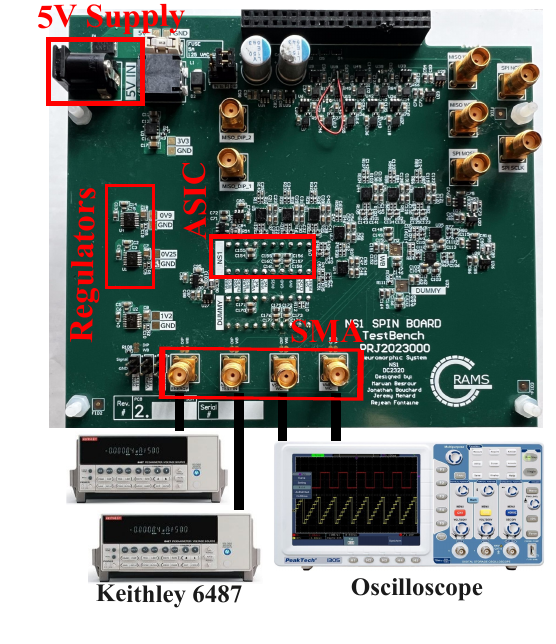}} 
\caption{Experimental Setup featuring the measurement equipment and the printed circuit board test bench custom designed for the chip.}
\label{fig:fig_setup1}
\end{figure}

Fig. \ref{fig:fig_setup1} shows the setup used to test and characterize the fabricated ASIC. The printed Circuit Board (PCB) includes voltage regulators to provide power supply to the ASIC and the interfacing circuits. The input of the analog spiking neuron circuit is tapped to a pad on the ASIC and the latter is connected to the outside via a SMA connector on the PCB. This was done to shield the input signals from external noise.

\begin{figure}[hptb]
\centerline{\includegraphics[width=0.9\linewidth]{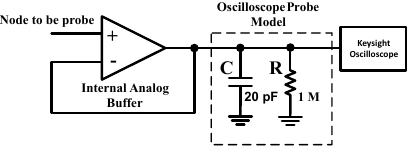}} 
\caption{Internal probing buffer for the output signals of the chip. Namely the membrane integration signal and the spiking signal.}
\label{fig:fig_setup2}
\end{figure}

There are also two internal analog operational amplifiers inside the chip for probing the spiking neuron's membrane signal and spiking output. Each buffer boosts its respective signal to make it observable on an external oscilloscope as Fig. \ref{fig:fig_setup2} shows. 

All the ASIC pins associated with the input, output, and power supply of the LIF circuit are connected to the benchmark equipment using SMA connectors on the PCB. The input and power supply are connected to a Keithley 6487 picoammeter and the outputs are connected to a Keithley Infinivision MSO3024A oscilloscope as depicted in Fig. \ref{fig:fig_setup1}.

\section{Methods For Characterization}

\subsection{ LIF Characteristics Measurement}

\subsubsection{Generation and integration of the LIF Synaptic Current} 

\begin{figure}[]
\centerline{\includegraphics[width=1\linewidth]{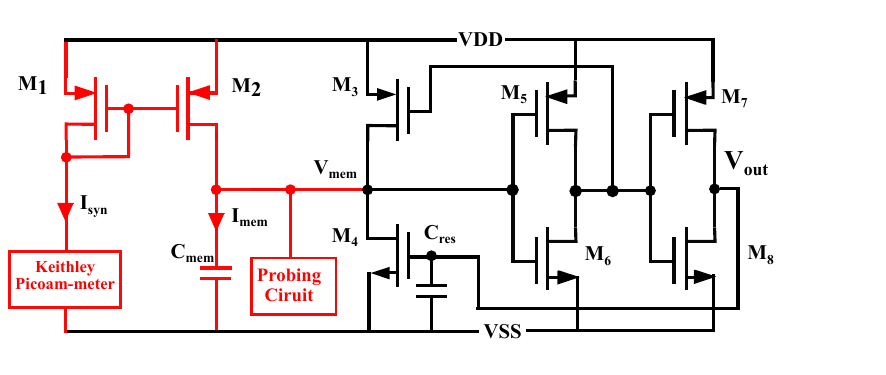}}
\caption{Simplified illustration of the experimental setup used to study the integration of synaptic current $I_{syn}$ in membrane capacitor $C_{mem}$ of the implemented analog LIF neuron.}
\label{fig:fig_metho_1}
\end{figure}

The incoming synaptic current is supplied by the picoammeter which generates a DC voltage between 0 V and 200 mV to the drain of $M_{1}$ as depicted in Fig. \ref{fig:fig_metho_1}, thereby setting the intensity of the input current to the LIF's membrane.  The power supply $VDD$ is set to 250 mV by a LT3045 low-dropout output (LDO) linear regulator with 500 mA driving capability, 0.8µVrms ultra-low noise (10 Hz to 100 kHz) and  76 dB  PSRR at 1 MHz, and an internal analog operational amplifier in buffer configuration allows the monitoring of the membrane voltage $V_{mem}$ via the oscilloscope using a low-impedance probe.

\subsubsection{LIF Spiking Behavior}

\begin{figure}[hptb]
\centerline{\includegraphics[width=1\linewidth]{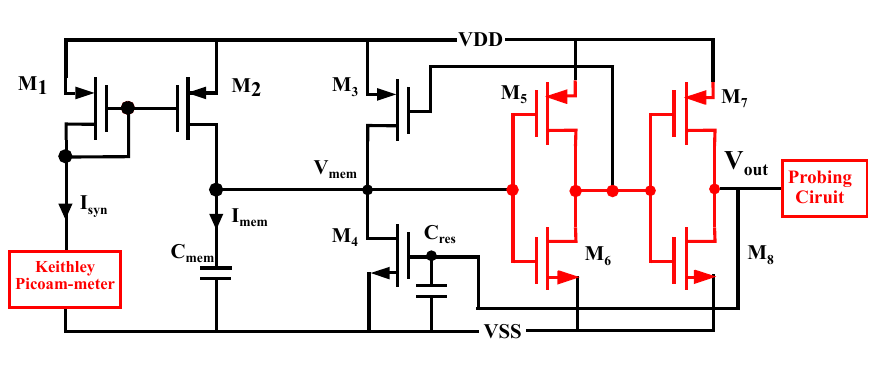}}
\caption{Experimental setup to study the spiking behavior of the implemented LIF neuron. The output waveform of the spiking stage is shaped by the neuron's intrinsic properties and the characteristics of the input synaptic current.}
\label{fig:fig_metho_2}
\end{figure}

The spike generation stage is validated by applying DC voltages at the drain and source of $M_{1}$ as shown in Fig. \ref{fig:fig_metho_2}, using the same set up as previously. An on-chip analog operational amplifier buffer configuration allows to observe the output voltage $V_{out}$ node with the oscilloscope. 

The very low voltage smooth pulses generated by the silicon neuron circuit are destined to be interfaced with digital circuits using a custom inverter designed with a low triggering threshold. This inverter will translate the pulses from the low-voltage power supply domain to the digital gates domain, ensuring compatibility with standard digital circuitry. The $1\; \mu s$ duration of the spike responses is maintained during this process, allowing precise temporal encoding to be preserved. Once translated, the pulses can be processed by subsequent digital circuitry for integration into larger-scale systems

The subsequent behavior of the leaky integrate-and-fire (LIF) neuron can be characterized by measuring the spiking frequency as a function of the input synaptic current 
and systematically analyzing the relationship between these values.

\subsubsection{Energy Consumption per Spike Event}

\begin{figure}[hptb]
\centerline{\includegraphics[width=1\linewidth]{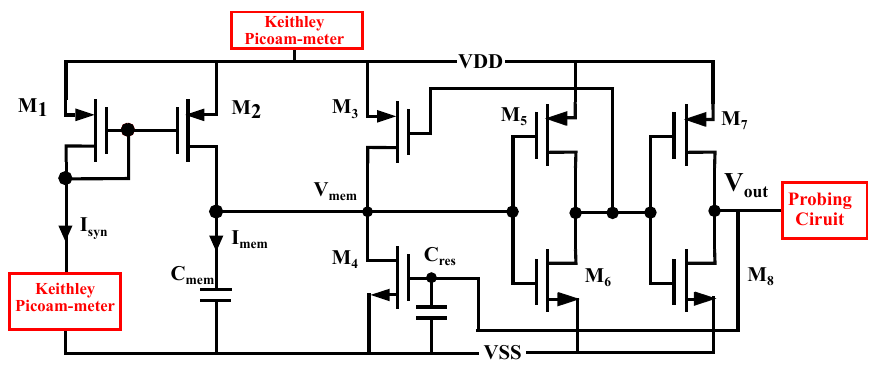}} 
\caption{Experimental setup to characterize the energy consumption by the implemented analog LIF neuron during spiking.}
\label{fig:fig_metho_3}
\end{figure}

This was achieved by sweeping the input voltage from 0 mV to 200 mV in 10 mV steps, using a one-second step and measuring the spiking frequency $Freq$ at the neural output with the oscilloscope and the sunk current $I_{rms}$ from the circuit with a second picoammeter plugged in series with the power supply node (see Fig. \ref{fig:fig_metho_3}).  This process is reiterated for each input current value, after which the root-mean-square (RMS) power consumption $P_{\text{RMS}}$ and energy per spike $E_{spike}$ are determined. The root mean square (RMS) power consumption is calculated as $P_{rms} = V_{rms} \times I_{rms}$ then the energy consumption per spike is computed as $E_{spike} =\frac{P_{rms}}{Freq}$. Using the proposed formulas, we may compute for each specific input current point, the power consumption and energy usage of the neuron circuit in Watts and Joules.

\subsection{Software Emulation}

To validate the scaling performance of the proposed LIF architecture in a neural network, a software model was developed with the \textit{PyTorch} machine learning library to simulate a network based on the measured LIF characteristics. This included defining the neural activation function based on the spiking frequency in response to the input synaptic current, measured across 20 samples of the same circuit to account for variability due to process and fabrication. The results were averaged to a single spiking function, which was then fitted using the \textit{curve fit} function from Scipy library \cite{SciPyCurveFit}. This provides a mathematical model of the activation function to implement a spiking neural network dynamic in \textit{PyTorch}.

A proof-of-concept SNN was built and trained with the MNIST dataset using half-precision arithmetic. Five different feed-forward networks were experimented with as shown in table \ref{tab:tab2}, where the first row indicates the topology used; for instance, [400-128-10] means the first layer has 400 neurons, the hidden layer has 128 neurons and the output layer has 10 neurons. 

To enhance the energy efficiency, we compressed the resolution of the MNIST dataset entries by shrinking the original 28x28 pixels image size (784 points total). Using the CV2 package and averaging each pixel with its neighboring pixels,  we reduced the input data size from 784 points to 400. 

The training of each network was executed on a GPU using the surrogate gradient technique, which enables to training of SNN by error backpropagation despite the  non-differentiability of these networks \cite{Zenke2018Surrogate, Shrestha2018SLAYER, Bellec2019Biologically}.

Our evaluation methodology also uses Post-Training Quantization (PTQ). The model is initially trained with half-precision floating-point weights and subsequently quantized to 4-bit resolution. We then incorporate the experimentally-derived activation function from chip measurements for accuracy evaluation. This approach simultaneously enhances computational efficiency and emulates the model's performance within a simulated resource-constrained environment.  Such emulation is crucial for assessing the architecture's real-time applicability and scalability for edge AI applications.


\paragraph{Surrogate Gradient}

The spike function of an SNN, which represent the action potential of a biological neuron, is inherently discontinuous \cite{Bellec2019Biologically,Shrestha2018SLAYER}, and it is typically modeled using the Heaviside function in computer simulations :

\begin{equation}
H(x) = 
  \begin{cases}
    1 & \text{if } x > V_{mem}^{th} \\
    0 & \text{otherwise}
  \end{cases}
  \label{equ_H}
\end{equation}

\noindent where $V_{mem}^{th}$ is the required membrane voltage for the neuron to fire. The firing pulse is obtained by resetting $V_{mem}$ after the function value changes to 1. In order to train the SNN using the popular error backpropagation algorithm, the derivative of the Heaviside function is needed. However, this derivative is the Dirac distribution, which is 0 everywhere except at the switching threshold where it tends to infinity. This prevents the use of regular gradient descent minimization as the gradient will either be zero or at saturation during computations. 
\cite{ZenkeMIT,zenke2017superspike,Zenke2018Surrogate}.

The surrogate gradient approach addresses the problem by keeping the Heaviside function during the forward pass, and computing its derivative with a continuous approximation during the backward pass. the \(Softmax\) function can be used for the purpose, with a parameter controlling its steepness near the Heaviside threshold. \cite{Zenke2018Surrogate}. The MNIST-trained SNN uses the measured f-I curves of the fabricated neuron as activation functions to closely emulate the circuit’s behavior. Surrogate gradient training was employed to address the non-differentiable nature of spike generation in the temporal domain. While rate-coding networks do not typically require surrogate gradients, this paper use of spike-latency coding for training necessitated this approach to enable efficient temporal learning.

The proposed LIF neuron circuit is intended as a scalable building block for neuromorphic systems leveraging efficient spiking neural network (SNN) architectures. While the MNIST emulation experiment using the the measured f-I curves employs a rate-coding approach as a proof-of-concept, the circuit is designed to support advanced coding schemes, such as spike-latency or event-based coding, which maximize the temporal precision of spiking neurons and minimize energy consumption by reducing the number of spikes required for computation.

\section{Results}

\begin{table*}[!ht]
\centering
\caption{Comparison of different neuron implementations in CMOS technology.}
\begin{adjustbox}{width=1\linewidth}
\begin{tabular}{|l|c|c|c|c|c|c|c|c|c|c|c|c|}
\hline
Parameters & \textbf{This work} & \begin{tabular}[c]{@{}c@{}} Indiveri \\ et al. 2006 \cite{indiveri_vlsi_2006} \end{tabular} & \begin{tabular}[c]{@{}c@{}} Joubert \\ et al. 2012  \cite{joubert_hardware_2012}\end{tabular} & \begin{tabular}[c]{@{}c@{}} Cruz-Albrecht \\ et al. 2012 \cite{cruz-albrecht_energy-efficient_2012} \end{tabular} & \begin{tabular}[c]{@{}c@{}} Wijekoon \\ et al. 2008 \cite{wijekoon_compact_2008} \end{tabular} & \begin{tabular}[c]{@{}c@{}}Danneville \\ et al. 2019 \cite{ DANNEVILLE201988}\end{tabular} & \begin{tabular}[c]{@{}c@{}}Sourikopoulos \\ et al. 2017 \cite{Sourikopoulos20174fj}\end{tabular} & \begin{tabular}[c]{@{}c@{}}Ferreira \\ 2019 \cite{Ferreira2019}\end{tabular} & \begin{tabular}[c]{@{}c@{}}Ronchini et al. \\ 2020 \cite{Ronchini2020TunableNeuron}\end{tabular} \\ \hline
Technology node & 28 nm & 0.35 um & 65 nm & 90 nm & 0.35 um & 65 nm & 65 nm & 55 nm & 180 nm \\ \hline
Supp. Volt. (V) & 0.25 & 3.3 V & 1 V & 1.2 V & 3.3 & 0.2 & 0.2 & (-0.1,0.1) & 0.2 \\ \hline
Implementation & Analog & Analog & Analog & Analog & Analog & Analog & Analog & Analog & Analog \\ \hline
Neuron type & LIF & LIF & LIF & LIF & Izhikevich & A-H & ML-based & ML-based & Izhikevich \\ \hline
Time scale & Accelerated & Biological & Accelerated & Biological & Accelerated & Biological & Biological & Accelerated & Biological \\ \hline
Mem. Cap. (F) & 3.47 fF & 432 fF & 500 fF & - & 100 fF & 5 fF & 4 fF & 9*934 aF & 35 fF \\ \hline
Energy (J/spike) & \textbf{1.61 fJ} & 900 pJ & 2 pJ & 400 fJ & 9 pJ & 2 fJ & 4 fJ & 3.6 fJ & 58.5 fJ \\ \hline
Spiking frequency & 300 kHz & 200 Hz & 1.9 MHz & 100 Hz & 1 MHz & 15.6 kHz & 26 kHz & 360 kHz & - \\ \hline
Area/neuron (\textmu m\textsuperscript{2}) & 34 & 2,573 & 120 & 442 & 2800 & 34 & 35 & 85 & 472 \\ \hline
\end{tabular}
\end{adjustbox}
\label{tab:tab1}
\end{table*}

We present here the measurement results of the implemented LIF neuron's characteristics with regard to the integration shape, the spiking behavior and frequency, the standby power, and the energy dissipation per spike. Beyond mere footprint reduction, our sub-2 fJ/spike performance arises from a combination of design and technology choices. First, by employing minimum-length channels in TSMC’s 28 nm node and biasing key transistors in deep sub-threshold, we realize picoampere-scale currents and negligible leakage, directly minimizing dynamic energy. Second, the use of an ultra-small membrane capacitance ($C_{mem}=3.47\ fF$) lowers the charge per spike ($ Q=C_{mem}V_{DD}$), shrinking both stored and discharged energy. Third, the neuron’s capability to fire at up to $300\ kHz$ means each high-power switching event occupies only a brief interval, and since $E_{spike}=P_{avg}/f$, a higher frequency directly drives down energy per event. Finally, operating at a supply voltage of only $V_{DD}=250\ mV$ exploits the quadratic dependence of dynamic power ($P_{dyn}\propto V_{DD}^2$) to curtail both static and dynamic dissipation. Together, these factors—sub-threshold transistor operation, minimal $C_{mem}$, high spiking rate, and low $V_{DD}$—enable our analog LIF neuron to achieve energy efficiency alongside its compact $34\ \mu m^2$ area.

\subsection{LIF Synaptic Current Generation}

\begin{figure*}[hptb]
\centering
\includegraphics[width=1\linewidth]{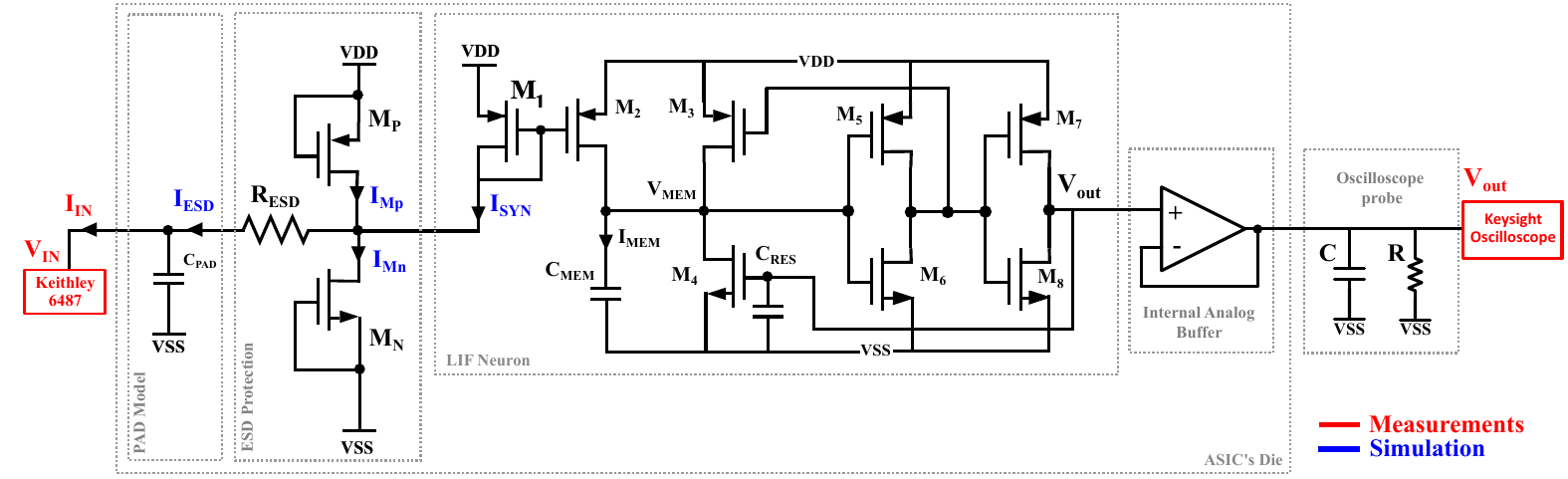}
\caption{Co-simulation and measurement approach to infer the LIF's input current ($I_{SYN}$). The measured values are displayed in red, and the post-layout simulated values are displayed in blue.}
\label{fig:fig_esd3}
\end{figure*}

The primary challenges in characterizing this architecture involve accurately measuring input current \(I_{IN}\) at picoampere levels. To achieve this, the methodology proposed in this paper is based on a combination of measurement and post-layout simulation to determine the input current. The input current to the LIF is sourced from a voltage source provided by the picoammeter connected to node \(V_{IN}\) in Fig. \ref{fig:fig_esd3}. The voltage is applied to the shorted drain and gate of transistor $M_{1}$ via a 200 ohm resistor (\(R_{ESD}\)) in the ESD protection circuit. In the next paragraph, \(I_{IN-sim}\) is the simulated \(I_{IN}\) and \(I_{IN-exp}\) is the measured \(I_{IN}\) using the experimental setup.

\begin{figure*}[hptb]
\centerline{\includegraphics[width=0.9\linewidth]{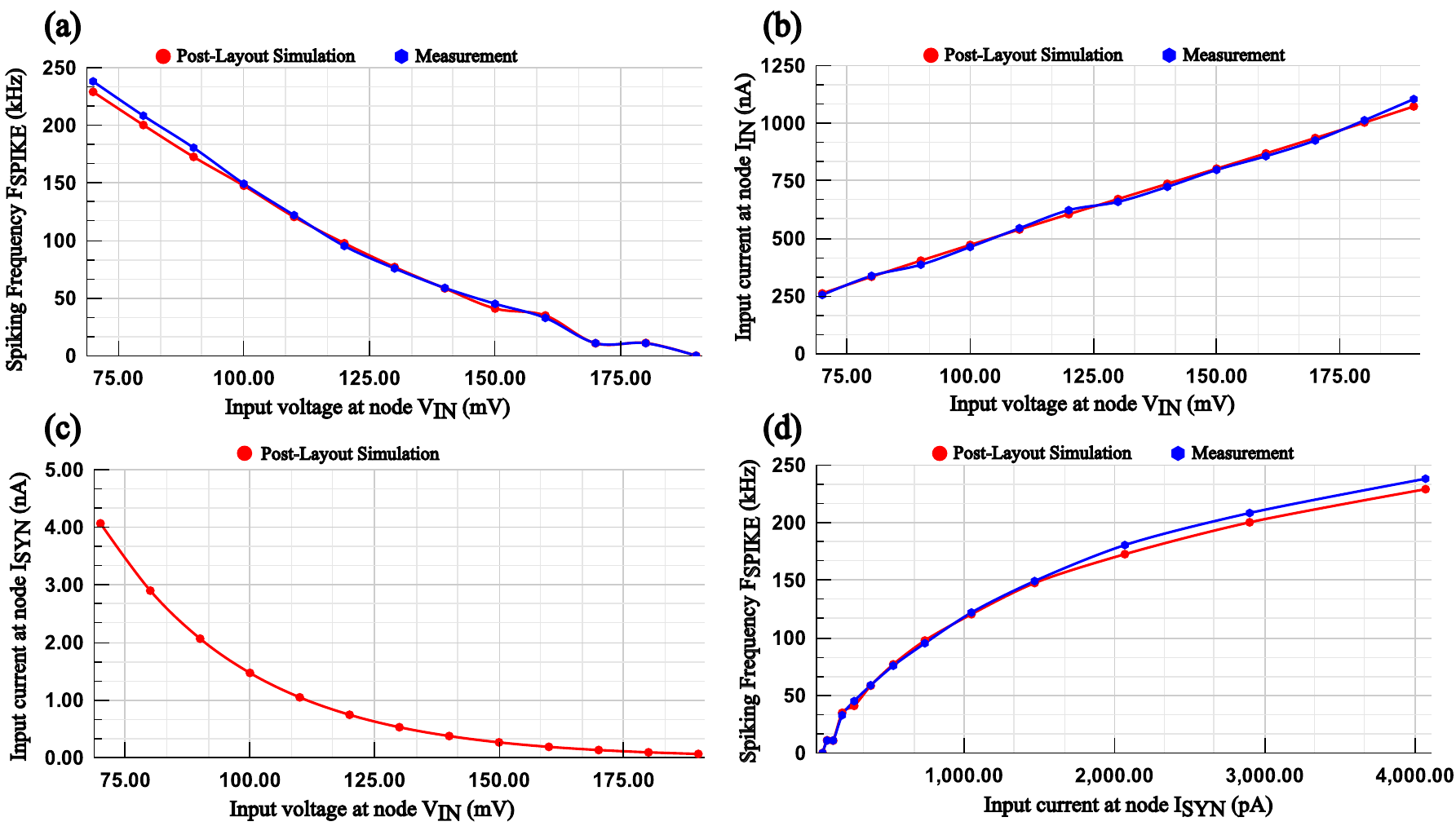}}
\caption{Quantitative analysis of the deviation between simulated and measured curves highlights the accuracy of the co-simulation measurement approach, further confirming its validity. (a) Depicts the spiking frequency of the LIF neuron at node $V_{out}$ in respect of the input voltage from the Keithley pico-ammeter $V_{IN}$. 
(b) Depicts the input current at node $I_{IN}$ depicted in Fig. \ref{fig:fig_esd3}  in respect of the input voltage from the Keithley pico-ammeter $V_{IN}$.
(c) Depicts the post-layout simulation results of the input synaptic current $I_{SYN}$ in respect of the input voltage from the Keithley pico-ammeter $V_{IN}$. 
(d) Depicts the spiking frequency of the LIF neuron at node $V_{out}$ in respect of the input synaptic current $I_{SYN}$.}

\label{fig:fig_cometho}
\end{figure*}

The first phase is to perform top-level post-layout simulations to determine the relationship between the tuple (\(I_{IN-sim}\), \(I_{SYN-sim}\)) and the resulting spiking frequency denoted \(F_{SPIKE-sim}\). Then, the input current  \(I_{IN-exp}\) is set with the Keithley picoammeter, and the resulting spiking frequency, \(F_{SPIKE-exp}\), is recorded with an oscilloscope connected to output node \(V_{OUT-exp}\).
The comparison of the post-layout simulation results with the measured current, which are depicted in figure \ref{fig:fig_cometho}, allows to estimate the input current at $M_{1}$'s drain \(I_{SYN-exp}\) for each measured spiking frequency  \(F_{SPIKE-exp}\). 

This method characterizes the relationship between the LIF neuron’s firing rate and its input synaptic current. Since no CMOS synapses are involved, the only relevant noise sources during current generation lie along the path from node $V_{IN}$ to node $V_{MEM}$ (Fig.~\ref{fig:fig_esd3}). Within this path, the principal contributors are the thermal and flicker noise of transistors M$_1$ and M$_2$, which form the current mirror. As these transistors integrate current onto $C_{MEM}$, their noise is likewise integrated and averaged over many spikes. Leakage currents at the gates, drains, and parasitic capacitances of M$_2$–M$_6$ contribute similarly minor noise. Although we cannot directly measure these picoampere-level noise contributions, post‐layout noise simulations confirm they do not materially affect performance. Suppressing such noise would require specialized processes (e.g., FDSOI) or cryogenic operation, which are beyond this work’s scope. Finally, spiking neural networks are inherently robust to input variability, so even with the inherent noise of the M$_1$–M$_2$ current mirror, the proposed design will scale effectively to large SNNs.

\subsection{Membrane Voltage Behavior over Time}

\begin{figure}[hptb]
\centerline{\includegraphics[width=1\linewidth]{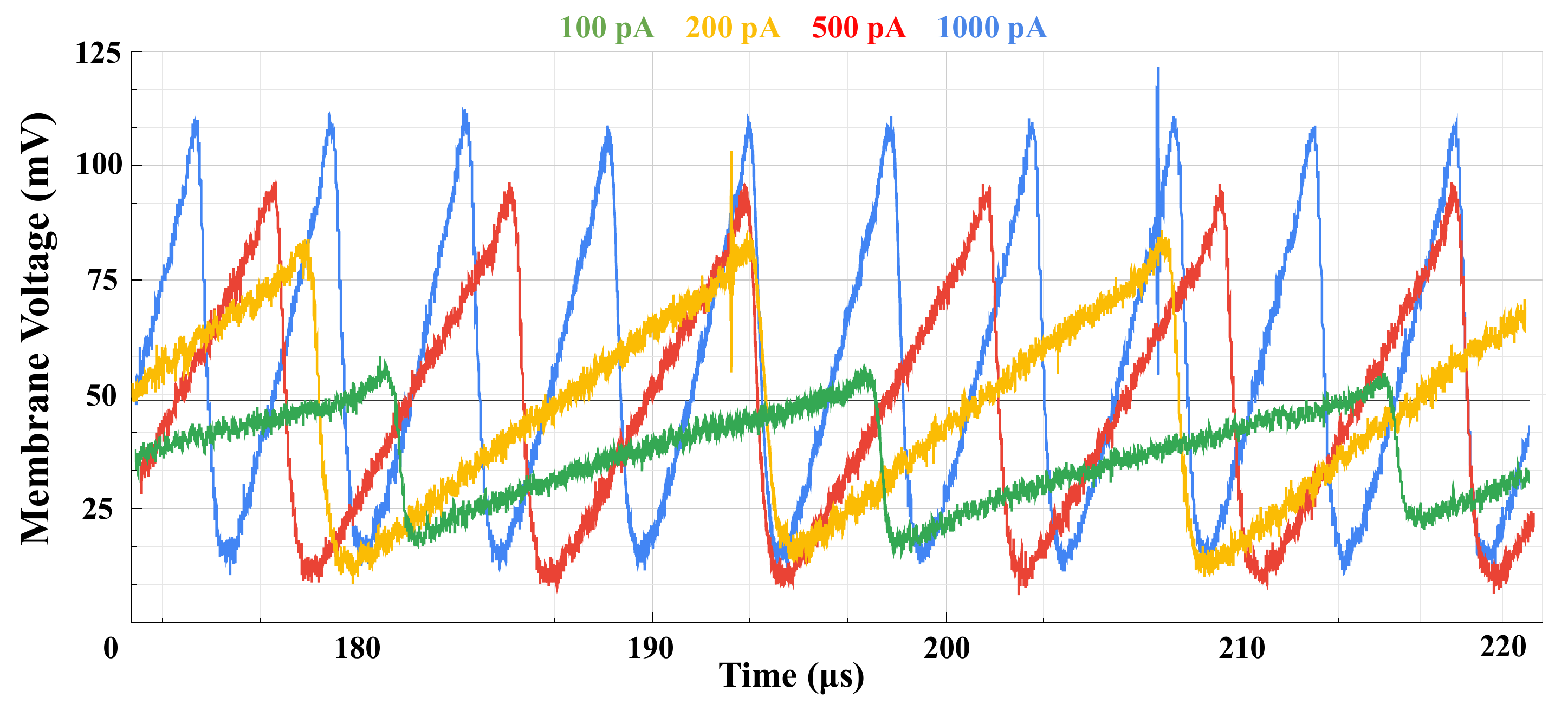}} 
\caption{Measured membrane voltage over time for four synaptic current values, showing the linear integration before spike emission.}
\label{fig:fig_vmem}
\end{figure}

Fig. \ref{fig:fig_vmem} illustrates the membrane voltage $V_{mem}$ response to different synaptic current levels in the proposed LIF neuron. Starting from the resting potential ($V_{reset} \approx 10$ mV), the voltage across the membrane capacitance $C_{mem}$ displays a consistent sawtooth pattern regardless of the input current magnitude. This pattern is characterized by a linear increase until the membrane voltage nears the threshold of the inverter pair $M_{5}$-$M_{6}$.

As $V_{mem}$ approaches this threshold, the activation of subsequent stages of the circuit gradually begins . This process eventually triggers transistor $M_{3}$ within the reset block, causing a sharp rise in the voltage across $M_{8}$ ($V_{out}$ up-swing), as detailed in \cite{Indiveri2011Neuromorphic}.

The variations in amplitude that lead to the voltage up-swing are due to the $M_{5}$-$M_{6}$ inverter operating in a sub-threshold mode \cite{5395622}. In this operational mode, the drain currents of transistors $M_{5}$ and $M_{6}$  increase exponentially with the increase in membrane voltage, which can be mathematically described by $V_{mem} = V_{GS6} = V_{SG5}$, following Eq. (\ref{eqn:eqVmem}). 

The exponential rise in current increases the charging rate of the gate capacitance of transistor $M_{3}$. This variable charging rate results in different activation timings for $M_{3}$, which in turn rapidly up-swings the membrane voltage $V_{mem}$. Consequently, $V_{mem}$ surpasses the inverter threshold $V^{M5-M6}_{INV}$, triggering a spike at the output. Following this spike, the rest of the circuit, particularly transistor $M_{4}$, resets $V{mem}$ back to $V_{reset}$.

The selection of the membrane capacitance $C_{mem}$ directly governs the integration rate and hence the neuron's spiking frequency. From Eq.\,(3), the integration slope is given by $dV_{mem}/dt = I_{mem}/C_{mem}$, so a smaller $C_{mem}$ produces a steeper voltage ramp, reducing the inter-spike interval and enabling higher maximum firing rates. Conversely, according to the energy relation $E_{spike}\approx\tfrac{1}{2}C_{mem}V_{DD}^2$, a larger capacitance increases the energy stored and dissipated per spike. Thus, $C_{mem}$ must be carefully sized to balance speed, energy consumption, and stability: excessively small capacitances yield faster spikes but amplify noise and process-variation sensitivity, while overly large capacitances improve robustness but incur higher energy costs and slower operation. In our design, $C_{mem}=3.47\,fF$ was chosen to support $>$ 300 kHz firing at $I_{syn}$ up to 3 nA, while keeping $E_{spike}$ below 2 fJ. 





Fig. \ref{fig:fig_results2} confirms these observations, displaying a plot of peak membrane voltages against the input current. Notably, all curves in the figure show a peak voltage of at least 60 mV, exceeding the sub-threshold swing of CMOS transistors to ensure an adequate dynamic range for the following inverter stage, which involves transistors $M_{5}$ and $M_{6}$ as shown in Fig. \ref{fig:fig_esd3}.

\subsection{Spiking Behavior over Time}

\begin{figure}[hptb]
\centerline{\includegraphics[width=1\linewidth]{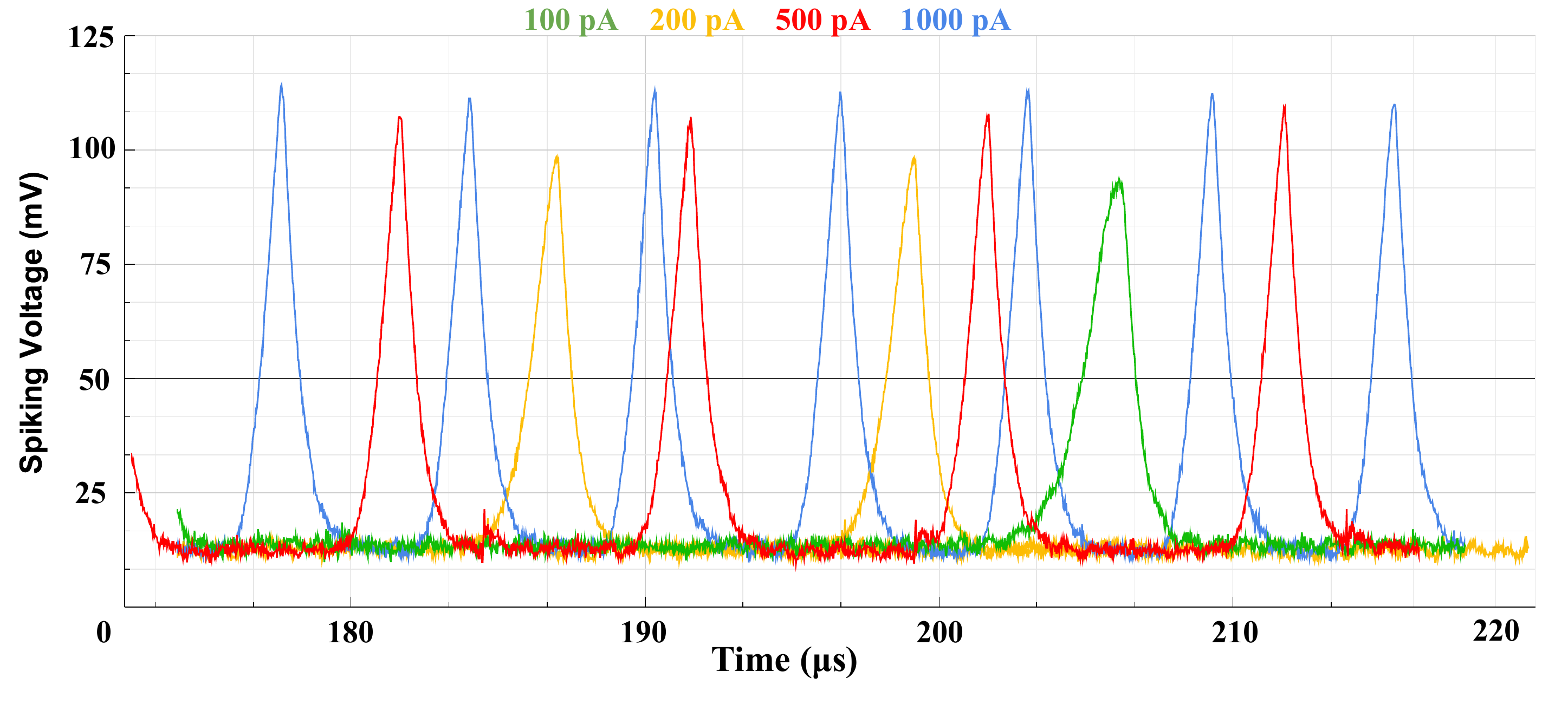}}
\caption{Measured periodicity of the generated spikes for four synaptic current levels. Each curve's color corresponds to the indicated synaptic current value.}
\label{fig:fig_results2}
\end{figure}

Fig. \ref{fig:fig_results2} illustrates the generated voltage spikes at different input synaptic current levels. The spike signals have a width of $1 \mu s $. 

The figure also shows a linear increase of the spike peak with the input current level as anticipated by equation Eq. \ref{eqn:eqVOH}. 
The reason is that given the technology node and the small transistor ratio's chosen, the speed of the reset block dominates the recursive link between the spike generation inverters and the reset transistors.
As varying the input current level leads to different integration slopes in capacitor $C_{mem}$, different delays occur before $V_{mem}$ reaches the switching threshold of inverter pair $M_{5}$-$M_{6}$. Then, $C_{mem}$ keeps charging and, after inverter pair $M_{7}$-$M_{8}$ also switches State, $C_{res}$ starts charging as well until it voltage reaches the threshold level to reset $C_{mem}$ through transistor pair $M_{5}$-$M_{6}$. Given the faster rise of $V_{mem}$ for larger input currents, a higher value of $V_{OH}$ across transistor $M_8$ is reached before $C_{mem}$ reset, hence the observed variation of the peak voltage with the intensity of the input current. 

\subsection{Spiking Frequency}

\begin{figure}[hptb]
\centerline{\includegraphics[width=1\linewidth]{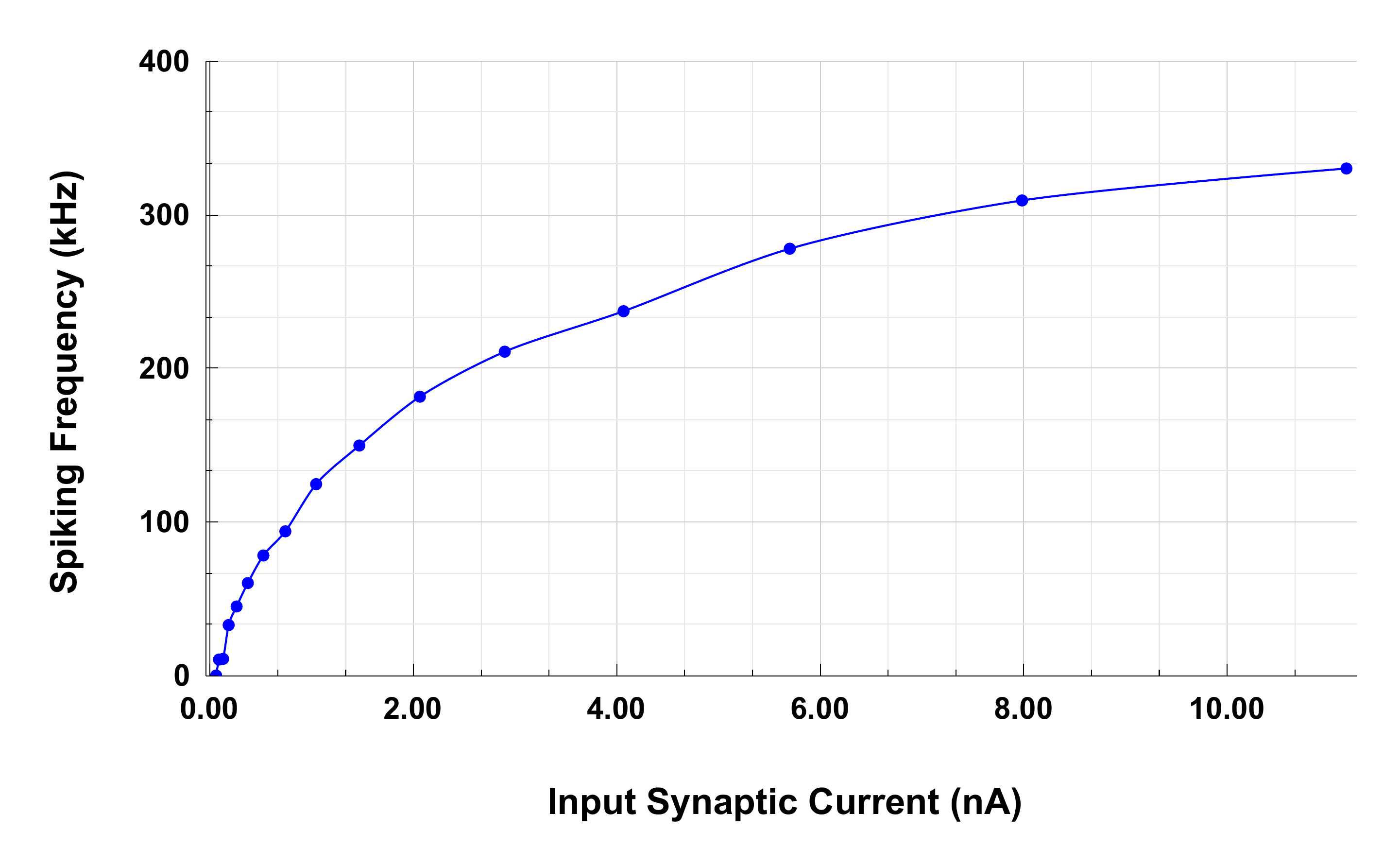}} 
\caption{Measured spiking frequency for input synaptic current levels from 0 pA to 10000 pA, showing the relationship between the spike generation rate and the intensity of synaptic current.}
\label{fig:fig_results_3}
\end{figure}

Fig. \ref{fig:fig_results_3} shows the firing frequency of the implemented LIF neuron for different input synaptic current amplitudes. The neuron's firing frequency increases with the input synaptic current, going from 10 kHz to 300 kHz in the 10 pA to 10000 pA input range, respectively. These results were collected from 20 ASIC samples randomly selected that were wire-bonded to a Dual-In-Line 20 (DIP20) package in order to test them using the PCB test bench.

\subsection{Energy Consumption per Spike Event}

\begin{figure}[hptb]
\centerline{\includegraphics[width=1\linewidth]{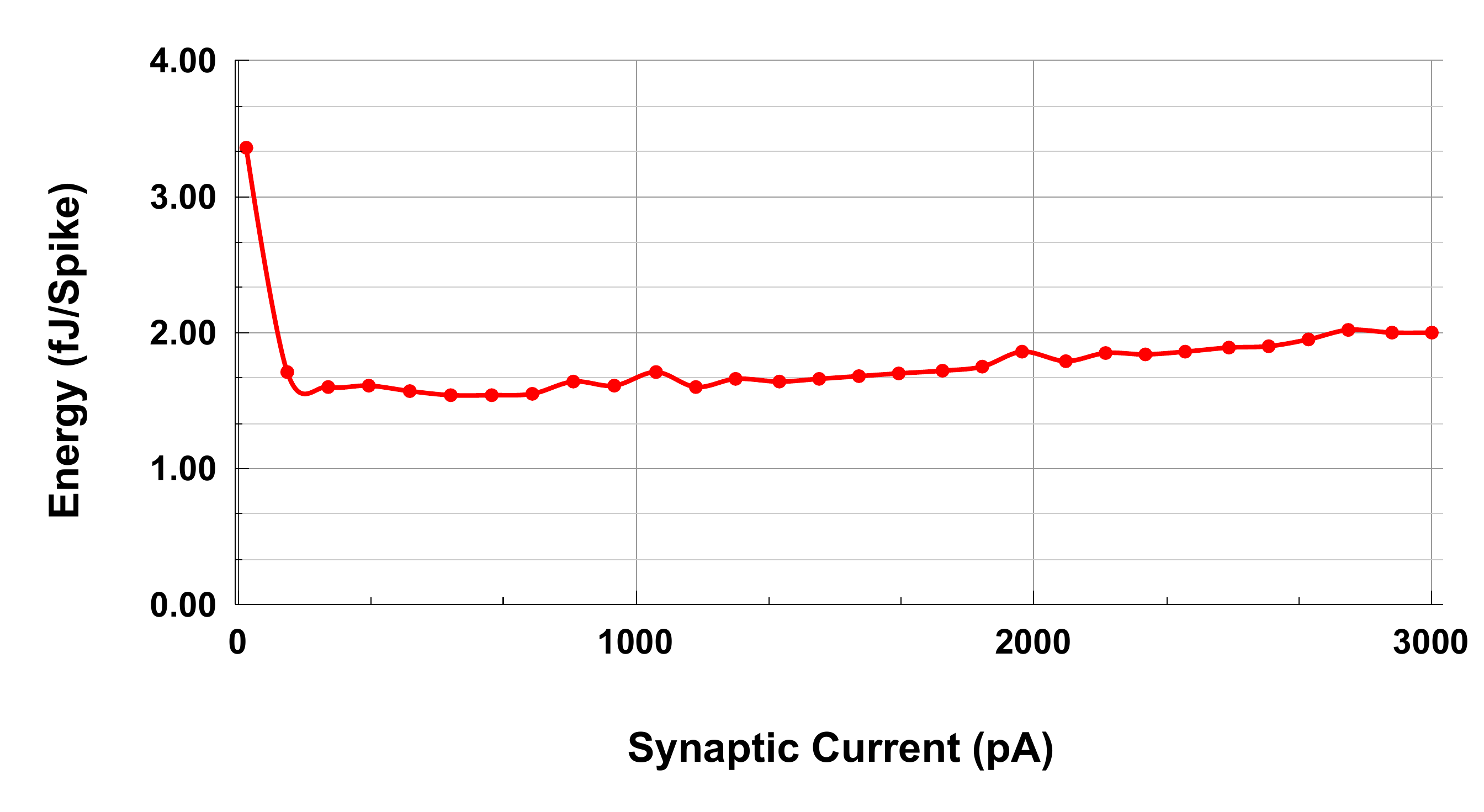}} 
\caption{Measured energy consumption per spike as a function of the input synaptic current.}
\label{fig:fig_results_4}
\end{figure}

Fig. \ref{fig:fig_results_4} shows the energy consumption per spike of the implemented LIF neuron for different input current values. The reported values account for both static and dynamic power consumption. Starting from about 3.3 fJ/spike corresponding to an input current of $22.6 pA$, the consumed energy converges down to a pseudo-plateau averaging 2 fJ/Spike. The minimum measured energy is 1.61 fJ/spike at an input current of $1500 pA$ 

\subsection{Software emulation of the characterized analog LIF}

Table \ref{tab:tab2} presented the results of evaluating the extracted biomimetic LIF model from our implemented analog spiking neurons to learn the MNIST dataset. The table showcases the test accuracy of different models both before and after PTQ, highlighting the biomimetic model, which achieved commendable accuracy even with low-precision parameters, indicating their suitability for efficient hardware adaptations.

\begin{table*}[!ht]
\centering
\caption{Accuracy performance of the model executed on GPU with 4-bit quantization}
\begin{tabular}{|l|c|c|c|c|c|}
\hline
Model Architecture & [400-128-10] & [784-128-10] & [784-256-10] & [784-128-10] & [784-128-10] \\ \hline
Learning rate & 0.0001 & 0.0001 & 0.0001 & 0.001 & 0.01 \\ \hline
Neurons in hidden layer(s) & [128] & [128] & [256] & [128] & [128] \\ \hline
Number of epochs & 20 & 20 & 40 & 20 & 20 \\ \hline
Batch size & 256 & 256 & 256 & 256 & 256 \\ \hline
Trained & (32-bit) & (32-bit) & (32-bit) & (32-bit) & (32-bit) \\ \hline
Quantized & (4-bit) & (4-bit) & (4-bit) & (4-bit) & (4-bit) \\ \hline
Test accuracy & 84.3\% & 80.9\% & 83.6\% & 86.6\% & 88.1\% \\ \hline
PTQ accuracy & \textbf{82.5}\% & 79\% & 80.6\% & 71.2\% & 67.5\% \\ \hline
Avg. Energy per Inference (pJ) & 483 & 753 & 917 & 679 & 693 \\ \hline
\end{tabular}
\label{tab:tab2}
\end{table*}

\section{Discussion}

\begin{figure}[h!]
\centerline{\includegraphics[width=1.2\linewidth]{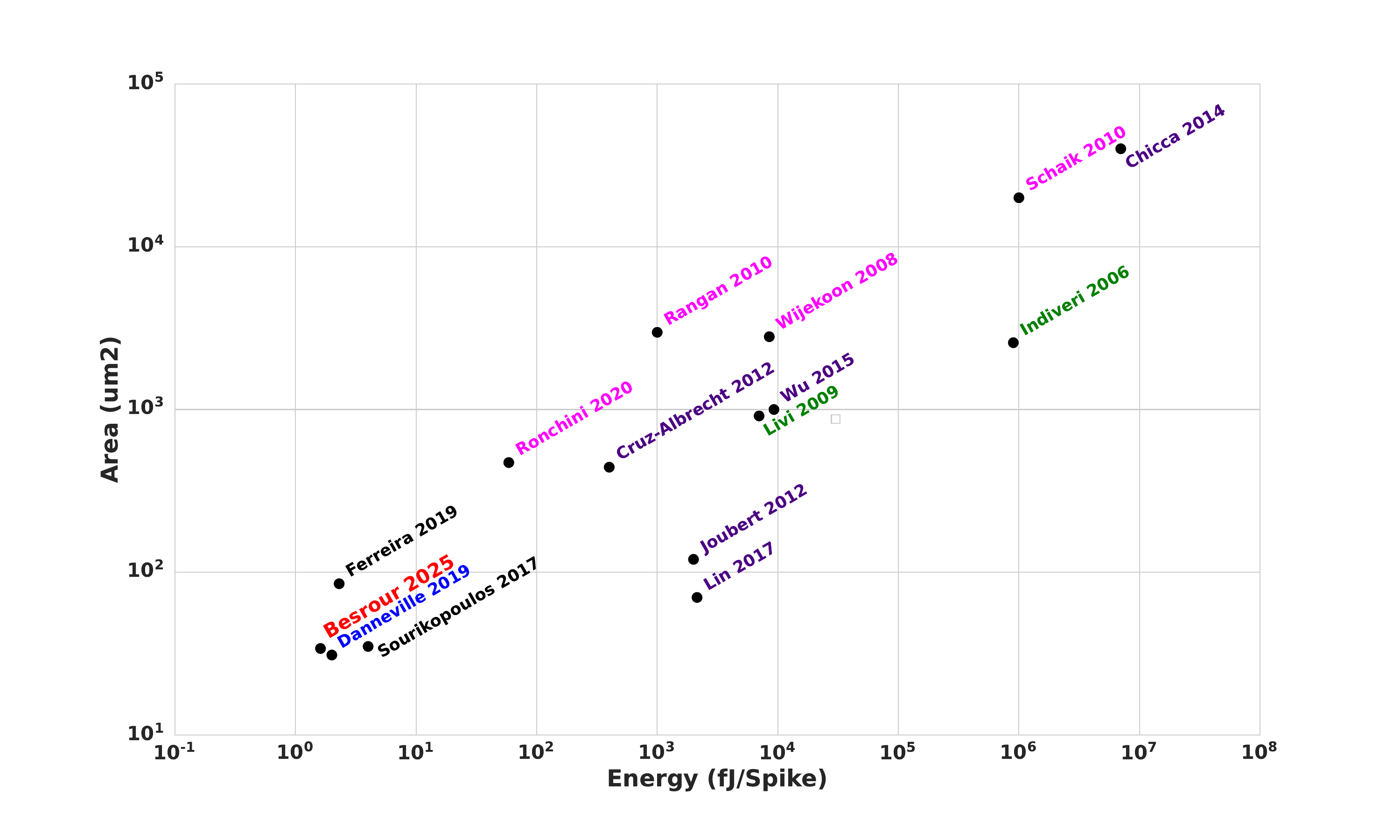}}
\caption{ Area vs. Energy of analog CMOS neurons color-classified by biological model. black is Moriss-Lecar, green is AdExif, blue is Axon-Hilcock, magenta is Izhikevich and indigo is LIF. This figure depicts a scatter of CMOS neuron models according to their active area versus dynamic energy consumption per spike. The ideal spot is at the origin where the energy and the silicon footprint are minimal.}
\label{fig:1}
\end{figure}

This paper presents an implementation of a spiking neuron in a 28 nm CMOS node. The proposed design has been fabricated and comprehensively characterized. Based on this detailed characterization, a model is developed using PyTorch to enable practical emulation on GPUs at a large neuron density. The comparison in Table \ref{tab:tab2} of our LIF design with published spiking neuron implementations in CMOS technology shows the achievement of our objective of a candidate for a large-scale neural network with minimal silicon footprint and energy consumption. This is shown clearly in Fig. \ref{fig:1} which provides a scatter diagram of recent spiking neuron implementations in CMOS technology, comparing their energy consumption per spike with their active area.

To the best of the authors knowledge, no prior work has combined operation at a 250 mV supply, a per-spike energy below 2 fJ, and an active area under 35 $\mu m^2$. Fabricated in TSMC 28 nm CMOS, the proposed neuron achieves 1.61 fJ/spike in only 34 $\mu m^2$ at 250 mV, representing an improvement in energy density over the closest analog designs and a reduction in active area compared to early 65 nm implementations, while still supporting spiking frequencies up to 300 kHz. This combination of scaling, sub-threshold integration, low-voltage operation and compact layout establishes a new performance benchmark for ultra-low-power, high-density analog LIF neurons. The results suggest that the proposed CMOS implementation of the LIF spiking neuron is a promising avenue for neuromorphic computing. A large-scale implemenations can be emulated by successfully deriving bio-mimetic models from analog neurons. This approach provides a robust method for verifying NeuroSOC designs before mass production and highlights the potential for creating efficient and scalable neuromorphic computing systems. The results of this paper showcase the effective implementation and characterization of the proposed LIF architecture, which accurately integrated input synaptic currents over time, resulting in a spiking behavior that depended on the synaptic current intensity in an increasing fashion. Notably, the energy consumption per spike event was found to depend in a pseudo-linear manner on the input current level. However, there are also inherent limitations of the proposed method. It is worth noting that the proposed method does not include the integration of synaptic connections, which are often considered a bottleneck in fully connected configurations. While memristor technology offers advantages such as a smaller active footprint compared to CMOS neurons, it still faces significant challenges in achieving the precision and reliability necessary for large-scale NeuroSoC implementations. The primary focus of this paper is to propose and validate a scalable LIF architecture specifically designed for CMOS implementation. The integration of synapses, while crucial for complete neuromorphic systems, falls outside the scope of this work and will be considered in future extensions. 

Fabricating an analog LIF neuron in TSMC’s 28 nm node posed several interrelated challenges. Operating key transistors in deep sub-threshold to achieve picoampere currents demanded careful matching addressed via common-centroid, multi-finger layouts. The 28 nm design rules yielded a layout mask that can inflate parasitic capacitances. To mitigate this, we performed extensive post-layout and corners simulations at FF through SS corners. Guided by these results, we tuned transistor aspect ratios and bias voltages to secure reliable firing behavior up to 300 kHz spiking and under 2 femto-Joules per spike. This methodology ensures a PVT reliable design that achieves its low-voltage, ultra-low-energy, and compact-area targets.

Furthermore, the energy consumption estimate for a full neural network inference is computed based on the experimental measurement per spike. To do this estimation, the pre-trained neural network is run on the test fraction of the dataset and the number of spikes produced by each neuron in every layer is determined to obtain the total number of generated spikes by the network during one inference. Then, this value is multiplied by the average measured energy per spike of 2 fJ/Spike. The recorded results are presented in Table \ref{tab:tab1}. This methodology offers valuable insights into emulating and forecasting the suitability of LIF neuron design for large-scale neuromorphic systems. While the energy consumption estimates presented in this work focus on the LIF neuron circuitry, it is important to note that synapses typically contribute significantly more to the total power consumption in neuromorphic systems, as observed in recent works \cite{Ferreira2019,DANNEVILLE201988} . Our estimates therefore provide an evaluation of the neuron’s energy efficiency in isolation, which was the primary focus of this study. Future work will aim to extend these analyses to include synaptic circuits to provide a more comprehensive understanding of the energy profile of the complete system.

The MNIST-trained neural network was included as a proof-of-concept to demonstrate the applicability of the proposed LIF neuron for practical spiking neural network (SNN) applications. The primary objective was to validate the feasibility of implementing such networks using the characteristics of the proposed silicon neuron. As this paper primarily focuses on hardware design and characterization, a detailed exploration of training metrics, such as learning rate, hyperparameters, accuracy over epochs, and dataset splits, was not the main scope of this work. These aspects will be investigated in greater depth in future studies focused on the deployment and optimization of SNNs for specific tasks. The observed performances figures are not intended to compete with state-of-the-art results but to demonstrate that the circuit can support spike-based computation in real-world tasks. The reduced accuracy is attributed to the simplified network architecture, lack of inhibitory mechanisms, and the use of experimentally measured circuit parameters, which were not fine-tuned for optimal performance

The energy-saving potential of spiking neural networks (SNNs) lies primarily in their ability to exploit efficient coding schemes, such as spike-latency or event-based coding, which minimize the number of spikes required for computation and take advantage of temporal sparsity. While SNNs can operate in a rate-coding context, this approach often negates many of the energy and computational benefits, as it requires temporal averaging to estimate rates and results in a linear relationship between spike count and accuracy. In contrast, binary or sparse coding schemes provide logarithmic scaling of spike count with accuracy, significantly improving efficiency. The MNIST experiment in this work employed a rate-coding approach for simplicity and proof-of-concept validation; however, the proposed neuron circuit is designed to support more efficient coding schemes in future implementations.

At the circuit level, the design departs from prior analog LIF implementations. First, the leak conductance is implemented implicitly via the sub-threshold operation of M$_3$–M$_4$, eliminating the need for a large on-chip resistor and reducing silicon area. Second, the integration block (M$_1$–M$_2$) operates at only 250 mV supply, extending sub-threshold current‐mirror operation into a voltage regime rarely explored by existing designs. Third, the reset mechanism employs capacitive coupling (C$_{res}$) and a threshold‐triggered transistor M$_4$ to achieve a precise 1 µs spike width without additional digital control. Fourth, the two‐stage inverter chain (M$_5$–M$_8$) is optimized for sub-threshold switching, using a low inversion threshold (appx. 60 mV) and minimal voltage swing to further cut dynamic energy. 




\section{Conclusion}

This paper presents an advancement in neuromorphic computing through the implementation and characterization of a Leaky Integrate-and-Fire neuron in 28 nm CMOS technology. The study's key achievement lies in a scalable, energy-efficient, and compact architecture suitable for large-scale neuromorphic systems. By successfully fabricating and characterizing the design, and emulating a neural network that uses a software model to fit written in PyTorch, the study demonstrates a practical use to build a neural network with it.

The research also underscores the potential of CMOS technology in neuromorphic computing by comparing various spiking neuron implementations, highlighting the trend toward more efficient designs. The proof-of-concept SNN, trained on the MNIST dataset, with reduced input data size and quantized weights, emphasizes the feasibility of employing biomimetic models for efficient hardware adaptations, even with low-precision parameters. Overall, the findings from this paper suggest that the proposed CMOS implementation of the LIF spiking neuron holds great promise for advancing neuromorphic computing, offering a viable pathway for creating efficient, scalable, and robust neuromorphic systems. The current design demonstrates the feasibility of hardware-based SNNs but is limited by the absence of inhibitory mechanisms and the precision of parameter control. Future work will focus on addressing these challenges to improve network performance and further validate the neuron design in more complex tasks.

Future work will focus on utilizing the proposed LIF neuron circuit in networks employing efficient coding schemes, such as spike-latency or event-based coding, to fully leverage the energy-saving and computational advantages of spiking neural networks.

\section*{Acknowledgments}
The authors would like to acknowledge and thank the Canadian Microsystems Corporation (CMC) and the Natural Sciences and Engineering Research Council (NSERC) for their financial support of this project and for providing the infrastructure essential for ASIC fabrication.



\bibliographystyle{ieeetr}
%

\bibliography{bare_jrnl_new_sample4.bib}

\newpage

\section{Biography Section}

\begin{IEEEbiography}
[{\includegraphics[width=1in,height=1.25in,clip,keepaspectratio,trim= 1cm 5cm 1cm 1cm]{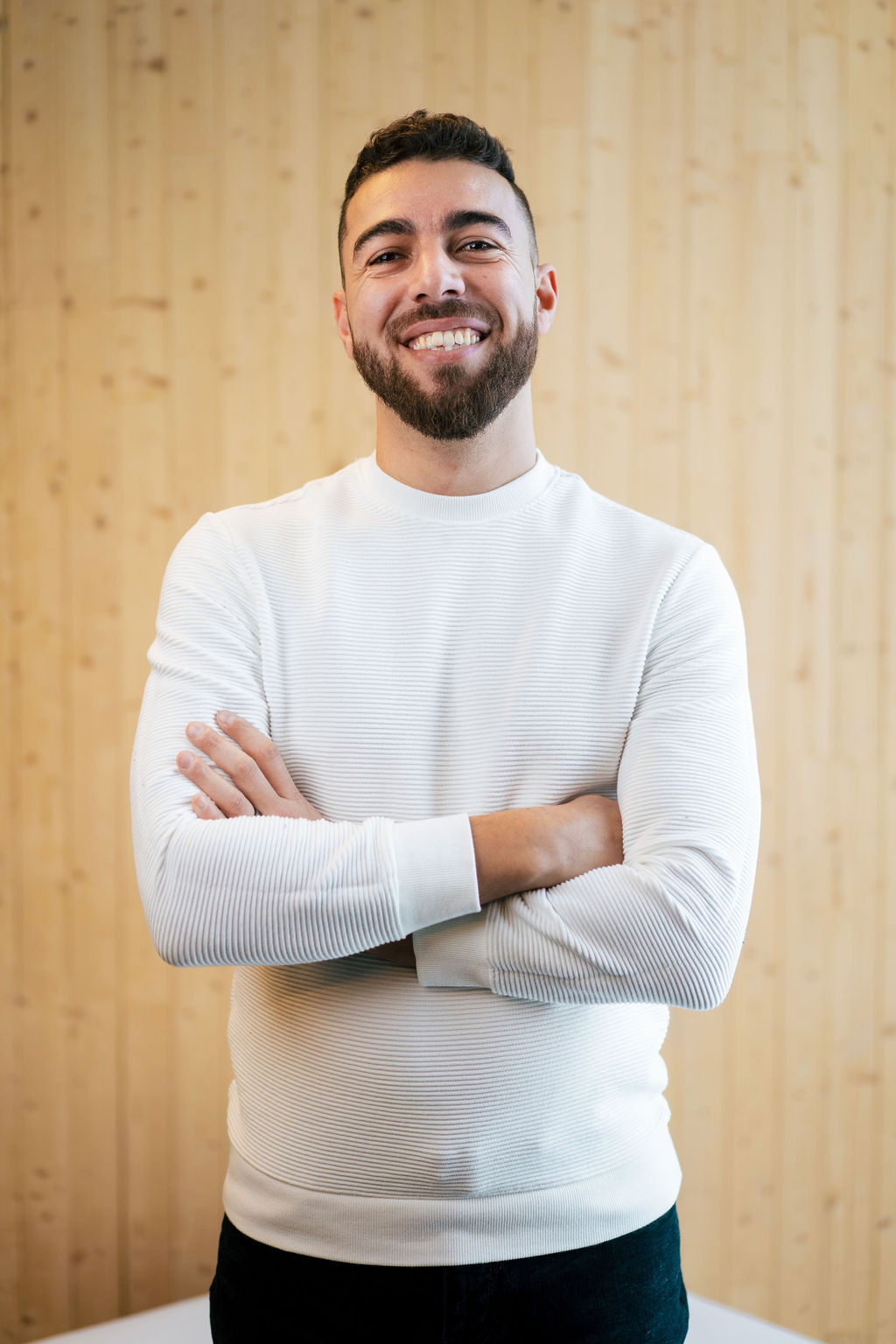}}]
{Marwan Besrour} received the bachelor’s degree
in electrical engineering with a major in analog
and mixed signals microelectronic from the École
Nationale d’Ingénieurs de Tunis (ENIT), Tunis,
Tunisia, in 2015, and the M.A.Sc. degree in electrical engineering from the Université Laval, Quebec,QC, Canada, in 2018. He is currently pursuing the Ph.D. degree in electrical engineering, with a focus
on custom neuromorphic processors for wireless
implantable medical devices, with the Université de
Sherbrooke, Sherbrooke, QC, Canada.
He joined the Groupe de Recherche en Appareillage Médical de Sherbrooke
GRAMS, Université de Sherbrooke, Sherbrooke, QC, Canada, in 2019, where
he designed custom ASICs for analog signal acquisition, electrical neurostimulation, optical data transmission, and RF communication. His current research
interests include analog and mixed-signal IC design, spiking neural networks,
and neuromorphic hardware.
\end{IEEEbiography}


\begin{IEEEbiography}[{\includegraphics[width=1in,height=1.25in,clip,keepaspectratio]{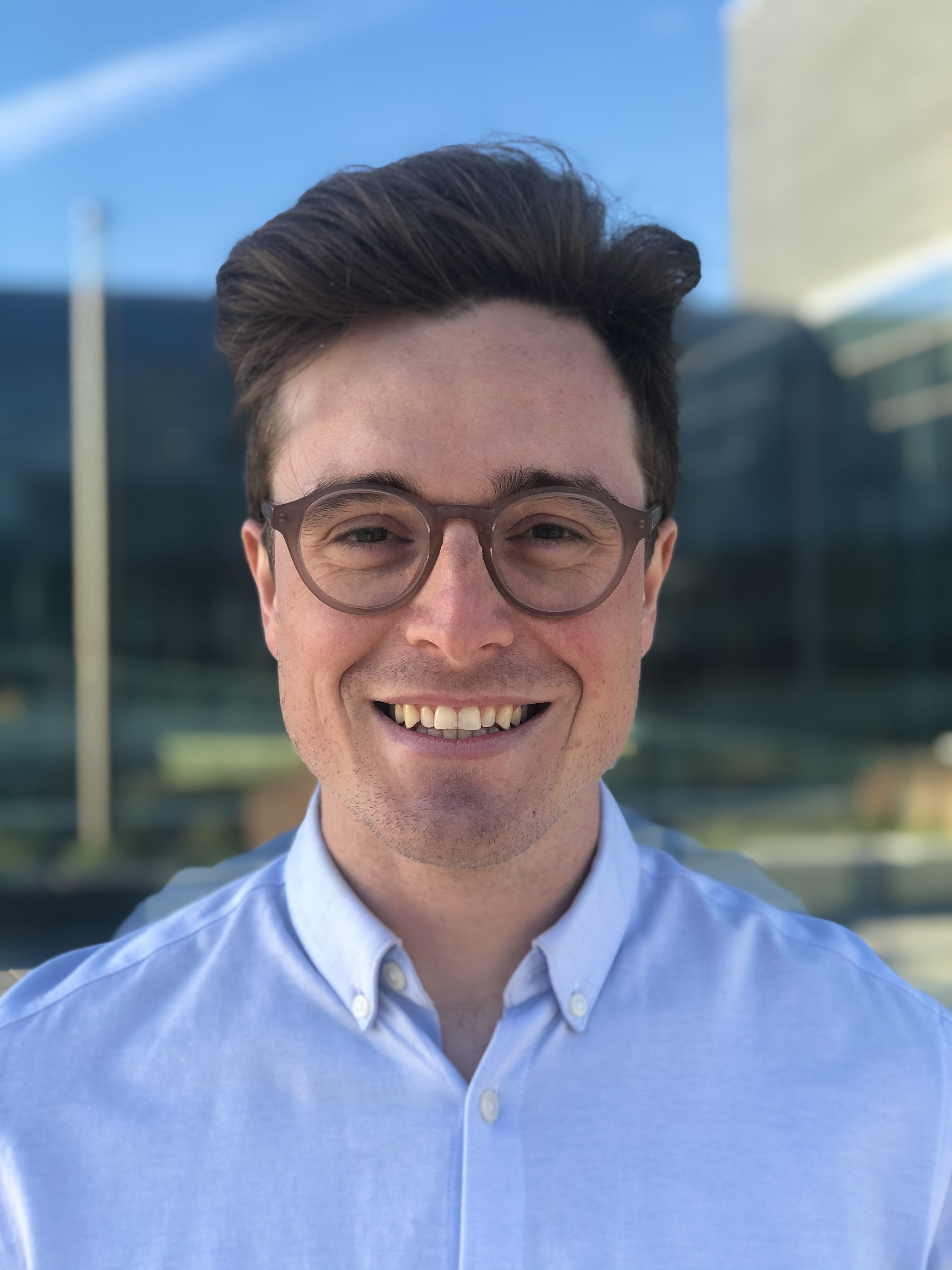}}]{Jacob Lavoie} 
received his B.Sc. in Neuroscience from Université de Montréal in 2016, his B.Eng. in Computer Engineering from Université de Sherbrooke in 2019, and his M.Sc. in Electrical Engineering from Université de Sherbrooke in 2020. He is currently an Applied Research Scientist at MILA. His research focus on artificial intelligence, embedded signal processing, reinforcement learning, and brain-computer interface.
\end{IEEEbiography}

\begin{IEEEbiography}[{\includegraphics[width=1in,height=1.25in,clip,keepaspectratio]{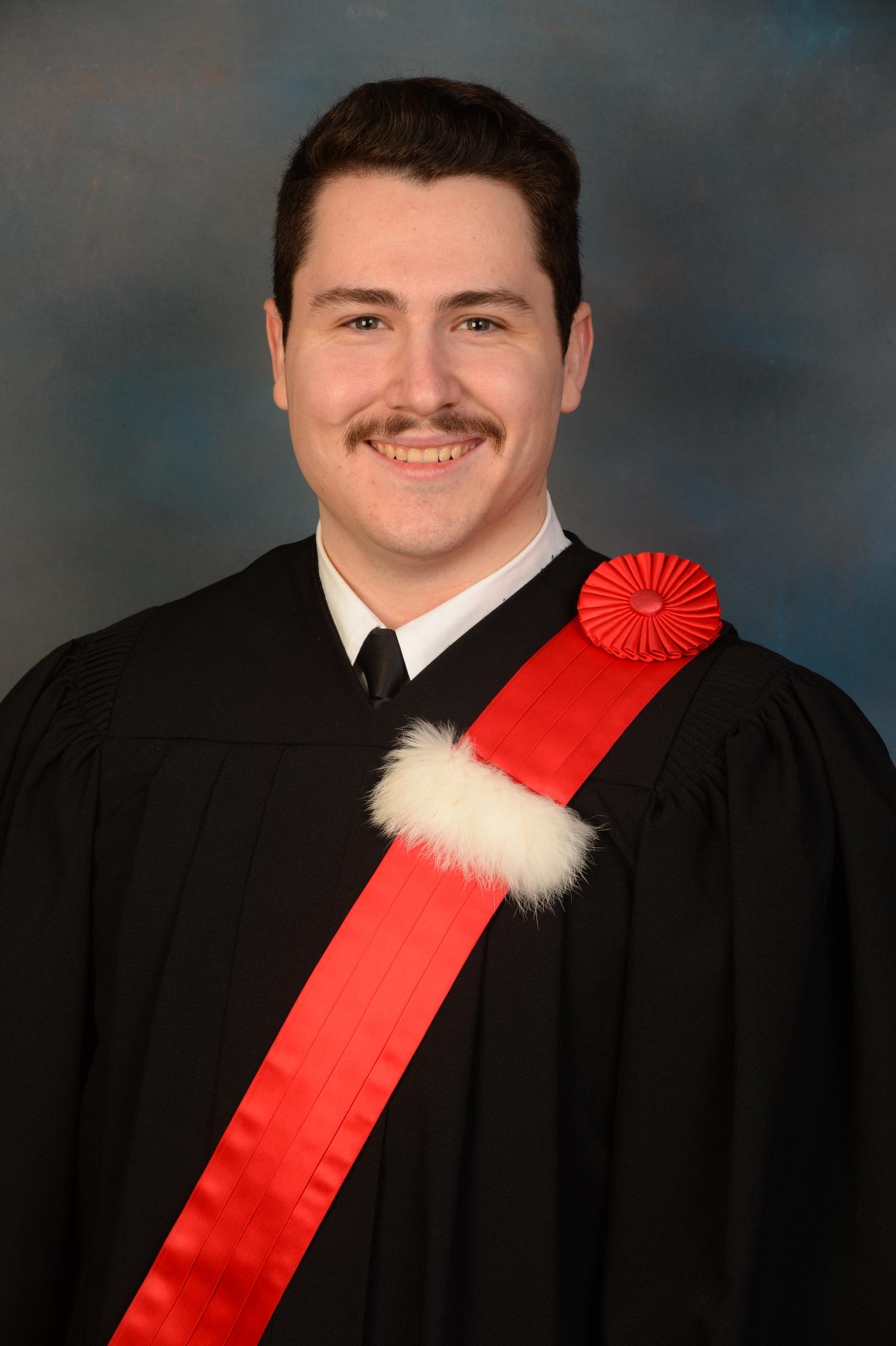}}]{Jérémy Ménard} received his bachelor's degree in electrical engineering from École de technologie supérieure (ÉTS), Montréal, Canada, in 2021. He is currently pursuing a Master of Applied Science (M.A.Sc.) in electronic design at Université de Sherbrooke, focusing on PCB design and wireless energy transfer for implantable medical devices. Since 2021, he has been a member of the Groupe de recherche en appareillage médicale de Sherbrooke (GRAMS), where he is involved in developing a wireless charging device and an external ultra-wideband (UWB) and Bluetooth acquisition unit designed to detect seizures up to 24 hours in advance.

\end{IEEEbiography}

\begin{IEEEbiography}[{\includegraphics[width=1in,height=1.25in,clip,keepaspectratio]{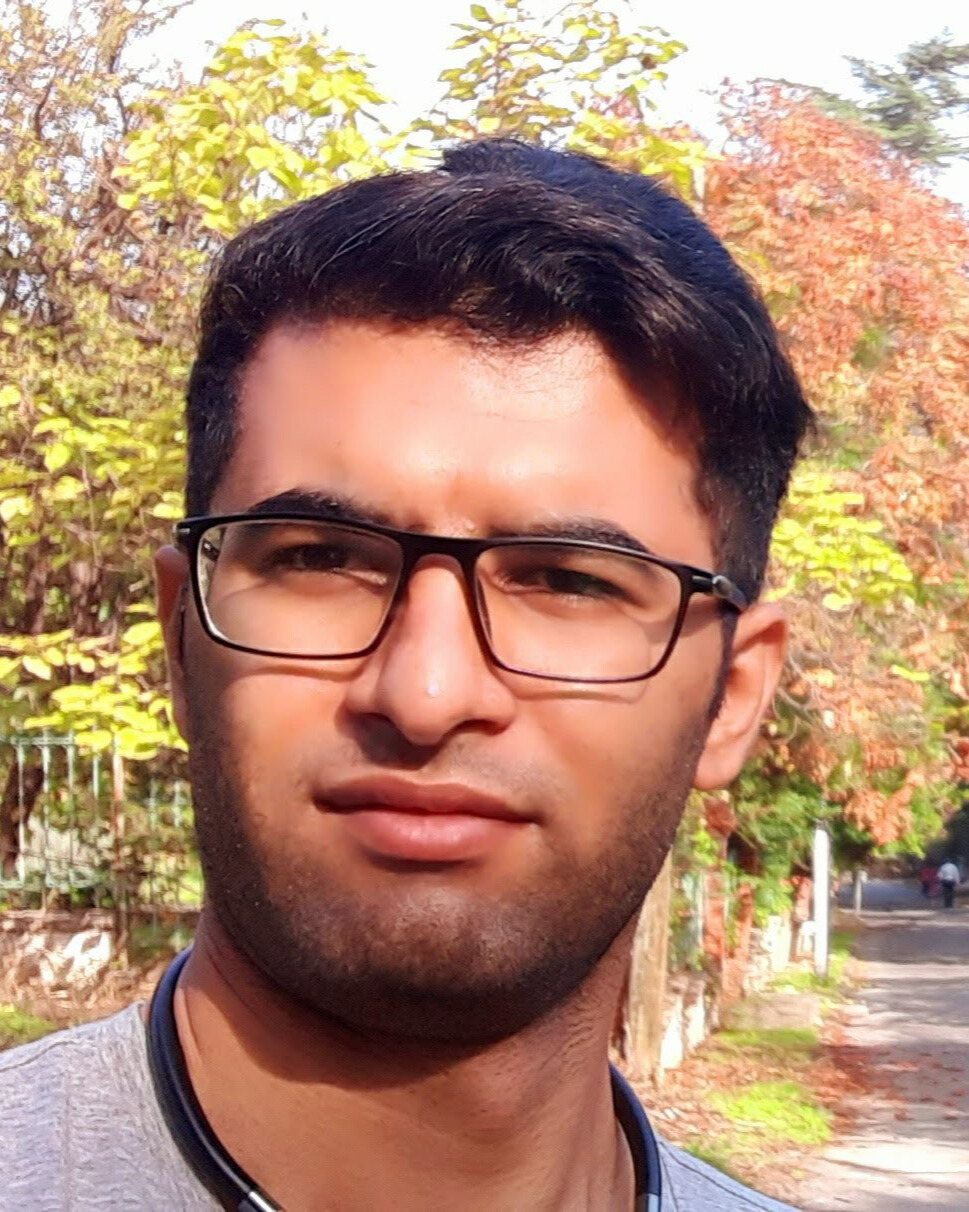}}]{Esmaeil Ranjbar Koleibi} received the bachelor’s degree M.Sc. degree in electrical engineering from Semnan University, Semnan, Iran, in 2014, and in 2017 respectively, with a focus on analog integrated circuits. He joined the Groupe de Recherche en Appareillage Médical de Sherbrooke (GRAMS), Université de Sherbrooke, Sherbrooke, QC, Canada, in 2021, where he is currently pursuing the Ph.D. degree in electrical engineering. His current research interests include UWB systems, wireless body area networks for implanted sensors. 
\end{IEEEbiography}

\begin{IEEEbiography}[{\includegraphics[width=1in,height=1.25in,clip,keepaspectratio]{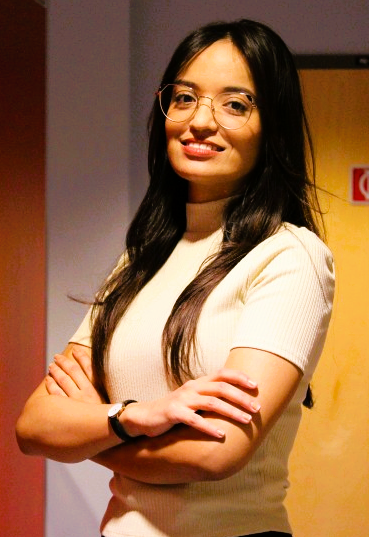}}]{Takwa Omrani} 
received the bachelor’s degree in electrical engineering with a major in analog and mixed signals microelectronic from the École Nationale d’Ingénieurs de Tunis (ENIT), Tunis, Tunisia, in 2021. She obtained her M.A.Sc. degree in electrical engineering from Université Paris Cité, Paris, France, in 2021, and another master’s degree from the École Nationale d'Ingénieurs, also in 2021. Her expertise lies in analog ASIC design, with a focus on high-speed communication system interfaces. She is currently serving as a Senior Chief Engineer for telecommunication and advanced computer systems in transportation industry.
\end{IEEEbiography}

\begin{IEEEbiography}[{\includegraphics[width=1in,height=1.25in,clip,keepaspectratio]{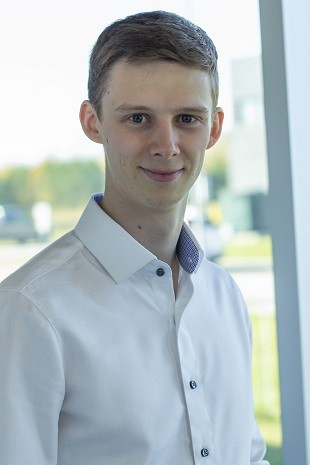}}]{Philippe Marcoux} received the bachelor’s degree in electrical engineering from the Université de Sherbrooke in 2019, and the M.A.Sc. degree in electrical engineering from the Université de Sherbrooke in 2022. He is currently working in research at Université de Sherbrooke designing custom electronics for different research projects. He joined the Groupe de Recherche en Appareillage Médical de Sherbrooke GRAMS,Université de Sherbrooke, Sherbrooke, QC, Canada, in 2018, where he design custom printed circuit boards for time-of-flight computed tomography.
\end{IEEEbiography}

\begin{IEEEbiography}[{\includegraphics[width=1in,height=1.25in,clip,keepaspectratio]{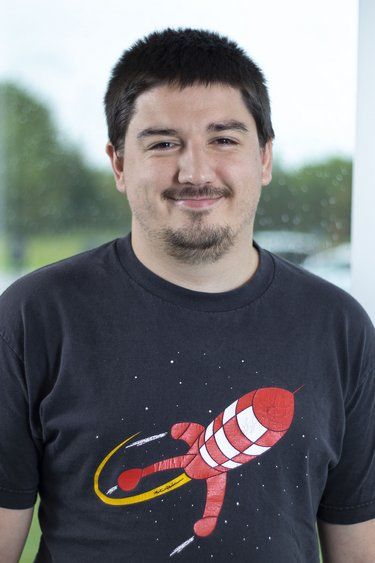}}]{Gabriel Martin-Hardy} 
received the bachelor’s degree in electrical engineering from the Université de Sherbrooke, Sherbrooke, QC, Canada, in 2018, where he is currently pursuing the M.A.Sc. degree in electrical engineering with the Groupe de Recherche en Appareillage Médical de Sherbrooke. He specialized in assembly processes for implantable medical devices. His current research interests include micro packaging, micro assembly process, nano assembly process and assembly characterization.
\end{IEEEbiography}

\begin{IEEEbiography}[{\includegraphics[width=1in,height=1.25in,clip,keepaspectratio]{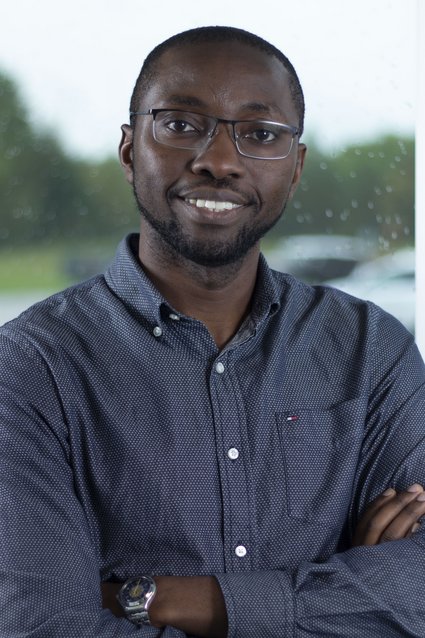}}]{Konin Koua} received the engineering degree from the Ecole Nationale Supérieure d'Ingénieurs de Caen, France in 2007, and the M.A.Sc. degree in electrical engineering from Université de Sherbrooke, Sherbrooke, QC, Canada in 2010. He held internship positions at Careland Information System, Shenzhen, Guangdong, China and at Brookhaven National Laboratory (BNL), Upton, NY, USA. Since 2010, he is a research professional at the Groupe de Recherche en Appareillage Médical de Sherbrooke (GRAMS), contributing and leading ASICs tape-outs on low-noise front-end electronics for positron emission tomography scanners, and on neural recording and stimulation (since 2017). In addition, he supports the development of new 2.5D assembly processes at the 3IT.micro prototyping platform, Université de Sherbrooke since 2014. In 2016, he was visiting scientist at the Stanford Linear Accelerator Center (SLAC), Menlo Park, CA, USA, to contribute to ASIC design for ATLAS experiment at CERN. Since 2014, he has held fixed-term consulting positions as layout designer at Cadence Design Systems, Montreal, QC, Canada, where he is still contributing to SerDes and memory interface IPs in advanced CMOS nodes (down to 5nm), as Layout Lead Design Engineer (From 2019). His current research interests include readout electronics for radiation instrumentation, wireless neural recording and stimulation and advanced assembly for 2.5D/3D integration.
\end{IEEEbiography}

\begin{IEEEbiography}[{\includegraphics[width=1in,height=1.25in,clip,keepaspectratio]{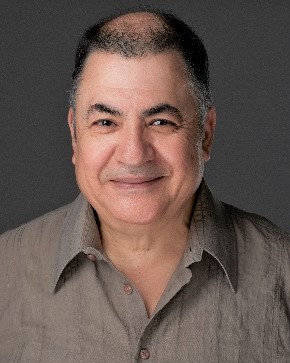}}]{Mounir Boukadoum} (Life Senior Member, IEEE) has a multidisciplinary background in physics and electrical and computer engineering. He received the Ph.D. degree in EE from the University of Houston, TX, USA, before joining the University of Quebec at Montreal (UQAM), Canada, where he is Professor of microelectronic engineering. He is also the current Director of the Quebec Strategic Alliance for Microsystems Research Center (ReSMIQ). His research interests include artificial intelligence in engineering analysis and design, with an emphasis on medical and biomedical applications. Prof. Boukadoum is an Active Member of IEEE, with involvement in four CASS conference steering committees and in the CASS Neural Systems and Applications Technical Committee. He is also a co-founder of the IEEE NEWCAS conference, now having interregional flagship status, of which he has chaired or co-chaired several editions. 
\end{IEEEbiography}

\begin{IEEEbiography}[{\includegraphics[width=1in,height=1.25in,clip,keepaspectratio]{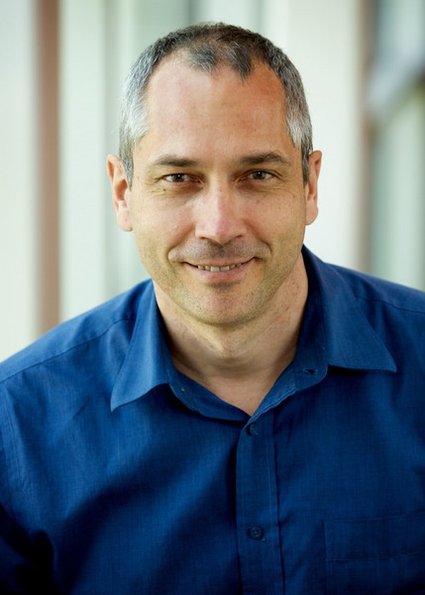}}]{Réjean Fontaine} received his bachelor and Ph.D. degrees at Université de Sherbrooke, Sherbrooke, Canada in 1991 and 1999. After spending a few years in the industry as an R\&D manager, he got back to academia in 2001 at Université de Sherbrooke where he is currently holding a full professor position in electrical engineering and computer engineering department. He successfully engineered two positron emission tomography scanners dedicated to small and mid-sized animal imaging. These works conducted to innovations related to electronic integration in the field recognized by many awards. He is the first awardee of the IEEE Emilio Gatti Radiation Instrumentation Technical Achievement Award (2019). He manages a Canada Research Chair Tier 1 dedicated to Time of Flight positron emission tomography and authored more than 250 scientific contributions including international journals, book and conferences. In 2012, he founded the 3IT.micro which he still manages. His current research interests include Time of flight PET and Time of Flight computed tomography (ToF-CT), a new research area his team introduced in 2018. 
\end{IEEEbiography}


\end{document}